\documentclass[journal]{IEEEtran}

\usepackage[utf8]{inputenc}
\usepackage[english]{babel}
\usepackage{lineno}
\usepackage{amssymb, amsmath, mathtools,bbm}
\usepackage{amsfonts,amssymb,amsthm,epsfig,url,epstopdf,array}
\usepackage{graphicx}
\usepackage{multirow}
\usepackage{hyperref}
\usepackage{enumerate}
\usepackage{polski}
\usepackage{color}
\usepackage[pdftex,dvipsnames]{xcolor}
\usepackage{float}
\usepackage{subcaption}
\usepackage{booktabs}
\usepackage{xargs}
\usepackage{pgf}
\usepackage{tikz}
\usetikzlibrary{arrows,automata,positioning,shapes.geometric}
\usepackage{relsize}
\usepackage{algorithm}
\usepackage{algpseudocode}
\usepackage{tablefootnote}
\usepackage{threeparttable}
\usepackage{cite}
\usepackage{ifpdf}
\usepackage{array}

\begin{document}

\title{Forward Composition Propagation for Explainable Neural Reasoning}

\author{Isel~Grau,~Gonzalo~N\'apoles,~Marilyn~Bello,~Yamisleydi~Salgueiro,~and~Agnieszka Jastrzebska
\thanks{Isel Grau, Eindhoven University of Technology, The Netherlands.}
\thanks{Gonzalo N\'apoles, Tilburg University, The Netherlands.}
\thanks{Marilyn Bello, Universidad de Granada, Spain.}
\thanks{Yamisleydi Salgueiro, Universidad de Talca, Chile.}
\thanks{Agnieszka Jastrzebska, Warsaw University of Technology, Poland.}
\thanks{Corresponding author: Isel Grau (i.d.c.grau.garcia@tue.nl)}
\thanks{The first and second authors contributed equally to the paper.}}

\maketitle

\begin{abstract}
This paper proposes an algorithm called Forward Composition Propagation (FCP) to explain the predictions of feed-forward neural networks operating on structured classification problems. In the proposed FCP algorithm, each neuron is described by a composition vector indicating the role of each problem feature in that neuron. Composition vectors are initialized using a given input instance and subsequently propagated through the whole network until reaching the output layer. The sign of each composition value indicates whether the corresponding feature excites or inhibits the neuron, while the absolute value quantifies its impact. The FCP algorithm is executed on a post-hoc basis, i.e., once the learning process is completed. Aiming to illustrate the FCP algorithm, this paper develops a case study concerning bias detection in a fairness problem in which the ground truth is known. The simulation results show that the composition values closely align with the expected behavior of protected features. The source code and supplementary material for this paper are available at \url{https://github.com/igraugar/fcp}.
\end{abstract}

\begin{IEEEkeywords}
explainable artificial intelligence, compositional explanations, deep neural networks, interpretable representations, fairness.
\end{IEEEkeywords}

\section{Introduction}
\label{sec:introduction}

In recent years, the performance of machine learning algorithms in solving complex tasks from a high volume of structured and unstructured data has caught the attention of industries, governments, and society. These techniques are increasingly deployed in specific decision-making tasks that directly impact everyone's daily life, such as medical diagnosis or treatment, recidivism prediction, personalized recommendations, hiring processes, etc. Therefore, there is a clear need to ensure the meaningful and responsible use of machine learning models. However, the best-performing machine learning techniques tend to build less transparent models, obstructing trust and raising questions about the accountability and fairness of the decisions.

Explainable artificial intelligence (XAI) studies measures and methods for obtaining interpretable models or generating explanations. When working with models that are not intrinsically interpretable, post-hoc methods are an alternative way of obtaining explanations that are understandable to humans. By and large, post-hoc explanation methods can either rely on the inner structures of the explained model or be completely model-agnostic. Prominent approaches in the latter category include feature attribution methods, such as Shapley additive explanations (SHAP) \cite{NIPS2017_7062}, feature importance \cite{Breiman2001,JMLR:v20:18-760}, local interpretable model-agnostic explanations (LIME) \cite{Ribeiro2016}, or global surrogate models \cite{grau2018interpretable,frosst2017distilling,GRAU20214919}. On the other hand, post-hoc explanation methods specifically designed for neural networks cover multilayer \cite{zilke2016deepred,pmlr-v70-shrikumar17a}, convolutional \cite{selvaraju2017grad,zeiler2011adaptive,bau2017network,Dabkowski}, graph \cite{Lyu2022,tan2022learning},  as well as recurrent \cite{arras2017explaining,choi2016retain} architectures.  

In particular, feature attribution methods designed for neural networks aim to determine the role of the problem features in the predictions produced by network systems. The majority of these approaches are backpropagation methods \cite{Simonyan14deepinside,10.1007/978-3-319-10590-1_53,DB15a,10.1371/journal.pone.0130140,pmlr-v70-shrikumar17a}, which means that they rely on the backward pass to propagate an importance signal through the layers. However, these methods compute the feature attribution for a specific prediction in the output layer only, and the majority of them neglect the negative influence of the features \cite{pmlr-v70-shrikumar17a}. Other studies \cite{bau2017network,mu2021compositional} focus on finding the composition of the neurons with regard to predefined concepts, including neurons in the hidden layers. However, the main limitation of these approaches is the need for an extra dataset annotated with predefined concepts when working with non-structured data.

This paper proposes a method termed Forward Composition Propagation (FCP) for computing compositions describing the neurons in a multilayered neural network. The compositions are expressed in terms of the features of a structured decision problem. Our algorithm produces a local explanation for a given instance after the neural network is trained. The negative or positive influence of the feature in the neural unit is represented by the sign of the composition. A case study in bias detection shows that the compositions capture the propagation of bias from the protected features to the output neurons through the entire network, helping illustrate the meaning of the composition vectors. Towards the end, this paper analyzes how different activation functions affect the propagation and, ultimately, the detection of the bias in the network.

The structure of this paper is organized as follows. Section~\ref{sec:literature} reviews the state-of-the-art methods on explainability for neural networks. Section~\ref{sec:method} presents the proposed FCP explanation method. Section~\ref{sec:simulations} illustrates the usability of FCP with a case study on fairness where the ground truth is known and presents feature-flipping experiments on benchmark datasets. Section~\ref{sec:remarks} concludes the paper and provides future research directions.

\section{Explainability methods for neural networks}
\label{sec:literature}

Deep Neural Networks (DNNs) have gained relevance in recent years due to their excellent performance in complex tasks, such as computer vision \cite{Liu2020a}, natural language processing \cite{Lauriola2021}, and recommender systems \cite{Napoles2020}. However, the features that allow these models to succeed in such application domains (multiple hidden layers, non-linear transformations, etc.) make it impossible for humans to follow or understand how the mapping from input data to predictions occurs. In other words, DNNs are black boxes limiting their deployment in real applications where trustworthy models are required \cite{Bai2021}.

The literature reports plenty of works that aim to provide explanations of neural network models. They can be broadly grouped into three categories: explanation by simplification, feature attribution explanation, and visual explanation \cite{BARREDOARRIETA202082}. Methods of the first and second categories are the most widely adopted to explain neural networks, while the third category is less common. The scarcity of the latter is due to the complexities of creating visualizations just from the inputs and outputs of a black-box model. In any case, as humans privilege the comprehension of visual data, visual explanation strategies are mainly applied in combination with feature attribution techniques.

Explanation by simplification focuses on, given a complex model, constructing a simpler surrogate model that behaves like its precursor but is easier to interpret. Decision trees and rules are the preferred surrogate models for simplifying neural networks \cite{Hailesilassie2016}. Early works in this category focused on networks with one hidden layer, e.g., DIFACONN-miner \cite{ozbakir2010soft}, KDRuleEx \cite{sethi2012kdruleex}, and CRED \cite{Sato2001}. They all extract decision trees, tables, or rules that describe the neural network based on the units of its hidden layer. More contemporary works such as DeepRed \cite{zilke2016deepred} extend these approaches to DNN by adding more decision trees and rules. In the literature reviewed, it is observed that as the complexity of the neural network increases, researchers prioritize the use of feature attribution methods. This is because simplifying a neural network with more hidden layers is a complex task.
  
Feature attribution methods aim to determine the role of the features in the predictions produced by the network. The majority of approaches are backpropagation methods, which means that they rely on the backward pass to propagate an importance signal through the layers. For example, in the context of image classification, the Gradient approach \cite{Simonyan14deepinside}, deconvolutional networks \cite{10.1007/978-3-319-10590-1_53}, and Guided Backpropagation \cite{DB15a} compute the gradient of the output neurons with respect to the input image, producing saliency maps. These approaches mainly differ in their way of handling the backpropagation of the gradient through non-linear units that transform negative activations into positive outputs, e.g., ReLU units. When backpropagating, discarding the negative gradients provokes that these approaches do not count for the negative influence of features \cite{pmlr-v70-shrikumar17a}. 

In contrast, Layerwise Relevance Propagation (LRP) \cite{10.1371/journal.pone.0130140,Montavon2019,lomazzi2023explainability} backpropagates a relevance score starting from the activation of the predicted class neuron, taking the input and their sign into consideration. Meanwhile, DeepLIFT \cite{pmlr-v70-shrikumar17a} compares the activation of each neuron to a reference activation and backpropagates this difference, allowing the information to propagate when the gradient is zero. However, neither LRP nor DeepLIFT allows associating neuron-level scores computed in intermediate (hidden) layers with problem features. Such an association is only possible once the scores have reached the input layer. Finally, deep Taylor decomposition \cite{MONTAVON2017211} considers each neuron as an object that can be decomposed and expanded, then aggregates and back-propagates these explanations from the output to the input layer. Backpropagating the importance signal limits these methods to obtain the feature attribution once the signal arrives at the input layer, ignoring the role of the features in the hidden layers.

Introduced by D. Bau et al. in \cite{bau2017network}, Network Dissection (ND) is an explanation method that quantifies the latent representations of convolutional neural networks using a data set of images with labels of concepts. The method considers each unit as a concept detector, giving a neuron-level understanding of the model. Thus, measuring the alignment between highly-activated units and concepts can determine whether the unit responds to the human-interpretable concept. Authors in \cite{mu2021compositional} present a generalization of the ND method termed Compositional Explanations (CE). Instead of using the basic concepts as in ND, the idea behind CE is to combinatorially expand the set of possible explanations under consideration with compositional operators on concepts. Then, using beam search, the authors generate complex compositional explanations matching the unit's activation values. This model-agnostic strategy was tested on computer vision and natural language inference, obtaining explanations with higher quality compared to ND. However, the main drawback of current compositional explanation methods is that they require a dataset with a dense inventory of predefined concepts. Therefore, if a concept is not in the inventory, the methods will not explain a neuron using those concepts.

\section{Forward composition propagation}
\label{sec:method}

This section presents the main contribution of our paper, which concerns a method ---termed Forward Composition Propagation (FCP)--- to explain the reasoning behind feed-forward neuronal networks. In short, our method describes each neuron with a composition vector expressed in terms of features describing the structured pattern classification or regression problem under analysis. Such composition vectors provide semantics to the hidden components of the network and are propagated through the whole network from the input layer to the last one.

Before presenting our explanation method, it seems to be convenient to formalize the reasoning mechanism of feed-forward neural networks. Let us assume a network with $N$ inputs, $H$ hidden layers with arbitrary width, and $M$ output neurons. Moreover, let $A^{(l)}$ be the activation vector associated with the $l$-th layer such that $a_{i}^{(l)} \in A^{(l)}$ denotes the activation value of the $i$-th neuron in that layer. In this formalization, $l=0$ is the input layer containing the feature values for a given instance, while $l=H+1$ is the last layer with the network's output. Equation \eqref{eq:reasoning} displays the reasoning mechanism used to compute the activation value of the $i$-th neural processing entity in the $(l+1)$ layer,

\begin{equation}
\label{eq:reasoning}
a_{i}^{(l+1)}=f\left(\sum_{a_{j}^{(l)} \in A^{(l)}} w_{ji}^{(l+1)} a_{j}^{(l)} + b_i^{(l+1)}\right)
\end{equation}

\noindent where $w_{ji}^{(l+1)}$ denotes the weights arriving at layer $l+1$, i.e., connecting the $j$-th neuron in the previous layer with the $i$-th neuron in the current layer, $b_i^{(l+1)}$ represents the bias weight attached to the $i$-th neuron in the $(l+1)$-th layer, while $f(\cdot)$ is an activation function that provides non-linearity to the network.

The FCP algorithm associates each neuron with an $N$-dimensional vector that expresses its composition in terms of the features describing the problem. These composition vectors are calculated for a given data point after the network has been trained. Let $\Theta_i^{(l)} = [ \vartheta_{i1}^{(l)},\ldots, \vartheta_{iN}^{(l)}]$ be the composition vector for the $i$-th neuron belonging to the $l$-th layer and $\Theta^{(l)}$ be the set of all composition vectors in that layer. In our algorithm, the composition vectors in the input layer will be vectors full of zeros; only the position $\vartheta_{ii}^{(l)}$ will take one to indicate that the $i$-th neuron is entirely described by the $i$-th problem feature. Equation \eqref{eq:fcp_raw} shows how to obtain the non-normalized $k$-th composition value for the $i$-th neuron in the following layer,

\begin{equation}
\label{eq:fcp_raw}
\Tilde{\vartheta}_{ik}^{(l+1)} =\sum_{j} w_{ji}^{(l+1)} \vartheta_{jk}^{(l)}|a_{j}^{(l)}|
\end{equation}

\noindent where $\vartheta_{jk}^{(l)}$ denotes the composition vectors carried over from the previous layer, $w_{ji}^{(l+1)}$ represents the weights connecting these two layers, and $|a_{j}^{(l)}|$ are the absolute activation values of neurons in the current layer. In our method, the neurons' activation values should be understood as the magnitude of their relevance in the network, regardless of whether their activation values are negative or positive. We consider this strategy to be a better approach to deal with negative activation values than zeroing out the negative values when using the ReLU activation functions, as observed in gradient-based methods such as \cite{springenberg2014striving,zeiler2013visualizing}.

The values of the compositions from Equation \eqref{eq:fcp_raw} are not bounded to an interval since they depend on the values of the weights connected to the $l+1$ layer. Therefore, each composition value is normalized by dividing it by the sum of the absolute composition values attached to that neural processing entity. Equation \eqref{eq:fcp_norm} shows the normalization process for the composition values,

\begin{equation}
\label{eq:fcp_norm}
\vartheta_{ik}^{(l+1)} = \frac{\Tilde{\vartheta}_{ik}^{(l+1)}}{\sum\limits_{v} |\Tilde{\vartheta}_{iv}^{(l+1)}|}
\end{equation}

\noindent where the composition values of the $i$-th neuron are indexed by $v$ in the denominator. The $l_1$ normalization, allows our explanations to fulfill that $\vartheta_{ik}^{(l)} \in [-1,1]$ and $\sum_i |\vartheta_{ik}^{(l)}|=1$ for any neuron, which facilitates their interpretation.

The pseudocode in Algorithm \ref{alg:pseudocode} summarizes the procedure for generating the compositions of an instance $x$. The algorithm requires as input a list of layers of the pre-trained neural network for which the forward pass on $x$ has been done already. The matrix of composition vectors for the input layer $\Theta^{(0)}$ is initialized with the identity matrix $I_N$. For each layer object, a method \textbf{activations()} returns the vector of activation values of all neurons in the layer and the method \textbf{weights()} returns the weight matrix associated with that layer. The \textbf{absolute()} and \textbf{sum()} methods perform their namesake operations on matrices. The $\otimes$ operator performs a column vector-matrix element-wise multiplication by operating the column vector with each column of a given matrix, provided that both the matrix and the vector have the same number of rows. The resulting product has the dimension of the matrix.

\begin{algorithm}
\caption{Forward Composition Propagation}
\label{alg:pseudocode}
\begin{algorithmic}

\Require $layers$

\State $\Theta^{(0)} \gets I_N$
\State $l \gets 1$
\While {$l <= H+1$} 
    \State $A^{(l-1)} \gets absolute(layers[l-1].activations())$
    \State $W^{(l)} \gets layers[l].weights()$
    \State $\tilde{\Theta}^{(l)} \gets (((A^{(l-1)})^{\top} \otimes \Theta^{(l-1)})^{\top} * W^{(l)})^{\top}$ 
    \For {all neurons $i$}
        \State $\Theta_i^{(l)} \gets \tilde{\Theta}_i^{(l)} / sum(absolute(\tilde{\Theta}_i^{(l)}))$
    \EndFor
    \State $l \gets l+1$
\EndWhile

\State \Return $\Theta$
\end{algorithmic}
\end{algorithm}

\begin{figure}[!htb]
\centering
\resizebox{\columnwidth}{!}{
    \includegraphics{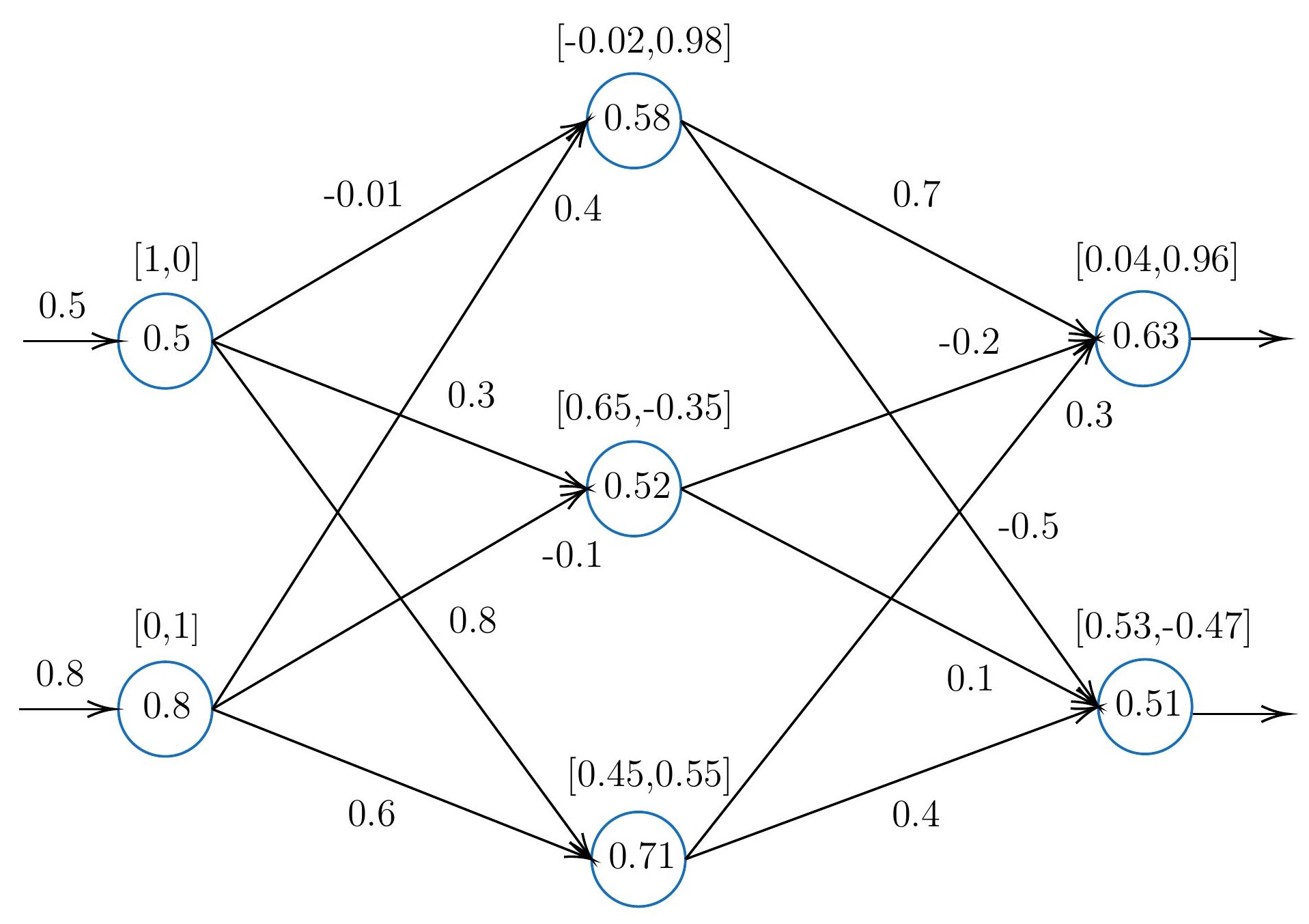}}
    \caption{Composition vectors computed by the proposed FCP algorithm for a simplified neural network having two inputs, a single hidden layer with three neurons without bias weights, and two output neurons. The numbers inside the neurons represent their activation values computed with a sigmoid transfer function, while the vectors on the top denote their associated composition vectors.} 
    \label{fig:example}
\end{figure}

Figure \ref{fig:example} depicts the composition vectors calculated by the FCP algorithm in a simplified feed-forward neural network composed of three layers. In this example, a sigmoid transfer function is used in all neurons while intentionally neglecting the bias weights for the sake of clarity. In the first layer, the matrix that results from stacking all composition vectors is the identity matrix. This matrix indicates which feature fully describes a given neuron in the input layer. In the upper layers, composition values are in the $[-1,1]$ interval, indicating the intensity and direction in which a certain feature affects the neuron. For example, the output neuron $o_1$ is described by the composition vector $[0.04, 0.96]$, which means that the second problem feature plays a larger positive role on that neuron when compared with the first one.

For more clarity, this section develops the execution of the pseudocode in Algorithm \ref{alg:pseudocode} for the simplified network in Figure \ref{fig:example}. The initialization step includes assigning the identity matrix to the composition matrix for the input layer $\Theta^{(0)}$, the input instance $[0.5,0.8]$ to the transposed activation vector ${A^{(0)}}^{\top}$, and the weights connecting the input layer and the hidden layer to $W^{(1)}$:

\begin{equation*}
\Theta^{(0)} =  \left[{\begin{array}{cc}
    1 & 0 \\
    0 & 1 \\
  \end{array} } \right] , {A^{(0)}}^{\top} = \left[{\begin{array}{c}
    0.5 \\
    0.8 \\
  \end{array} } \right] ,
\end{equation*}

\begin{equation*}
  \text{and } W^{(1)} =  \left[{\begin{array}{ccc}
    -0.01 & 0.3 & 0.8 \\
    0.4 & -0.1 & 0.6 \\
  \end{array} } \right] .
\end{equation*}

In the first step, the activation vector for the input layer is $\otimes$-operated with the compositions for the same layer and the obtained matrix is transposed, resulting in a temporal matrix given below:  

\begin{equation*}
({A^{(0)}}^{\top} \otimes \Theta^{(0)})^{\top} = \left[{\begin{array}{cc}
    0.5 & 0 \\
    0 & 0.8 \\
  \end{array} } \right] = R^{(0)} .
\end{equation*}

In the second step, the $R^{(0)}$ matrix is multiplied with the weight matrix $W^{(1)}$, using the common matrix multiplication operator and the result is transposed, leading to the matrix of raw composition values:

\begin{equation*}
(R^{(0)} * W^{(1)})^{\top} = \left[{\begin{array}{cc}
    -0.005 & 0.32 \\
    0.15 & -0.08 \\
    0.4 & 0.48 \\
  \end{array} } \right] = \tilde{\Theta}^{(1)} .
\end{equation*}

The third step of the algorithm consists of a row-wise normalization of the composition values while keeping their sign, resulting in the composition matrix $\Theta^{(1)}$ for the hidden layer: 

\begin{equation*}
\Theta^{(1)} = \left[{\begin{array}{cc}
    -0.02 & 0.98 \\
    0.65 & -0.35 \\
    0.45 & 0.55 \\
  \end{array} } \right] .
\end{equation*}

Now, the previous three steps are repeated to compute the composition matrix of the output layer. Similarly, the activation column vector of the hidden layer ${A^{(1)}}^{\top}$ and the weight matrix $W^{(2)}$ connecting the hidden layer with the output layer are initialized:

\begin{equation*}
{A^{(1)}}^{\top} = \left[{\begin{array}{c}
    0.58 \\
    0.52 \\
    0.71 \\
\end{array} } \right] \text{and } W^{(2)} =  \left[{\begin{array}{cc}
    0.7 & -0.5 \\
    -0.2 & 0.1 \\
    0.3 & 0.4 \\
\end{array} } \right] .
\end{equation*}

The first step $\otimes$-operates the activation values and the compositions of the hidden layer and transposes the product, resulting in the matrix: 

\begin{equation*}
({A^{(1)}}^{\top} \otimes \Theta^{(1)})^{\top} = \left[{\begin{array}{ccc}
    -0.012 & 0.338 & 0.319 \\
    0.568 & -0.18 & 0.391 \\
  \end{array} } \right] = R^{(1)} .
\end{equation*}

The second step obtains the raw composition values for the output layer:

\begin{equation*}
(R^{(1)} * W^{(2)})^{\top} = \left[{\begin{array}{cc}
    0.0201 & 0.5514 \\
    0.5338 & -0.466 \\
  \end{array} } \right] = \tilde{\Theta}^{(2)} .
\end{equation*}

And finally, the composition values are normalized for each output neuron in the layer:

\begin{equation*}
\Theta^{(2)} = \left[{\begin{array}{cc}
    0.04 & 0.96 \\
    0.53 & -0.47 \\
  \end{array} } \right] ,
\end{equation*}

\noindent resulting the in values shown in Figure \ref{fig:example}.

Overall, composition vectors computed by the FCP algorithm allow for explaining how the predictions were obtained using the network's inner knowledge structures and understanding the role of each feature in these predictions. This happens because the hidden neurons are no longer black-box entities bearing no meaning when it comes to the problem domain. It is worth noting that the bias weights are not considered when computing the composition vectors since they do not map to any explicit feature. Therefore, using a bias regularizer in the training phase to keep these weights as small as possible is advised to generate more reliable explanations. 

\section{Numerical simulations}
\label{sec:simulations}

This section illustrates the results of the proposed FCP algorithm through a case study concerning fairness in machine learning. Here, we compare the rankings of the compositions extracted by FCP with the relevance values computed by LRP and the feature attribution values extracted by the SHAP method. Toward the end of the section, we extend the experimentation to several datasets showing the effect on performance when removing the most important features according to our method.

\subsection{Methodology and data for the case study}
\label{sec:simulations:methodology}

The FCP algorithm is tested using a fully-connected neural network with four layers: an input layer, two hidden layers, and an output layer. The two hidden layers have $2N$ and $N$ hidden neurons, respectively, with $N$ being the number of problem features. These neural units will operate with either Exponential Linear Unit (ELU), Leaky Rectified Linear Unit (Leaky ReLU), Sigmoid, or Hyperbolic Tangent transfer functions. The categorical cross-entropy is adopted as the loss function, while output neurons are equipped with Softmax transfer functions. The weights associated with the multi-layer networks are computed using the well-known Adam optimization algorithm \cite{kingma2017adam} with the initial learning rate set to 0.001, the number of epochs set to 100, and the batch size set to 32. After using a stratified split of $80\%$ of the dataset to build the model and $20\%$ for testing purposes, this network configuration results in a 0.785 accuracy. In this experiment, the need for hyper-parameter tuning or cross-validation is discarded since the model already achieves the baseline accuracy reported in the literature and the generalization of the underlying model is out of the scope of the experiment. To validate the correctness of the FCP algorithm, this experiment assesses the extent to which the composition vectors align with the role of protected features observed in the literature for the case study. Protected features refer to personal characteristics such as race or ethnic origin, gender, religion or belief, disability, age, or sexual orientation, for which a person belonging to a minority group might be discriminated against.

The selected case study is the German Credit dataset \cite{dua2017uci}, which contains the demographic characteristics of clients of a German bank. This dataset is used for classifying loan applicants as good or bad credit risks. Applicants are described by the following 20 features: \textit{age}, \textit{credit amount}, \textit{credit history}, \textit{months of the credit}, whether the person is a \textit{foreign worker}, \textit{housing status}, \textit{installment rate}, \textit{job status}, \textit{existing credits}, \textit{people liable}, \textit{other debtors}, \textit{other installments}, \textit{gender}, \textit{employment since}, \textit{residence since}, \textit{property ownership}, \textit{purpose of credit}, \textit{savings account status}, \textit{checking account status}, and \textit{telephone}. The dataset contains the information of 1000 loan applicants, with 700 belonging to the good class and 300 to the bad class. Based on the literature, the features \textit{age} and \textit{gender} are considered protected features \cite{bellamy2018ai,friedler2019comparative,delobelle2021ethical}. It should also be mentioned that gender/female and age/young (younger than 25 years old) are deemed protected groups for this dataset. The preprocessing steps included:
\begin{enumerate}[(i)]
    \item normalization of numeric features using the max-min scaler
    \item encoding target classes as integers, and
    \item re-coding the nominal protected feature \textit{sex$\&$marital status} to include only \textit{gender}-related information.
\end{enumerate}

When it comes to the protected features, it has been reported in the literature that there is bias against young people, while men are often favored \cite{10.1145/3457607,10.1145/3368089.3409704,10.1145/3338906.3338937}. In a nutshell, the bias patterns in the dataset indicate that males and older applicants are more likely to receive good credit compared to females and younger counterparts. In the case of the numerical feature \textit{age}, it would suffice to verify that there is a positive (negative) correlation between age and the corresponding composition values for the case of good (bad) credit. In the case of the nominal feature \textit{gender}, the predictions computed by the neural network are contrasted with the predictions computed directly from the composition vectors associated with the decision neurons. In other words, the decision class will be given by the neuron with the largest composition value for the gender feature. As a result, it is expected to observe a close resemblance between the predictions produced by the neural network and the ones obtained from the composition values.

\subsection{Results and discussion for the case study}
\label{sec:simulations:results}

\begin{figure*}[!htb]
\center
    \begin{subfigure}{0.24\textwidth}
	\center
	\includegraphics[width=\textwidth]{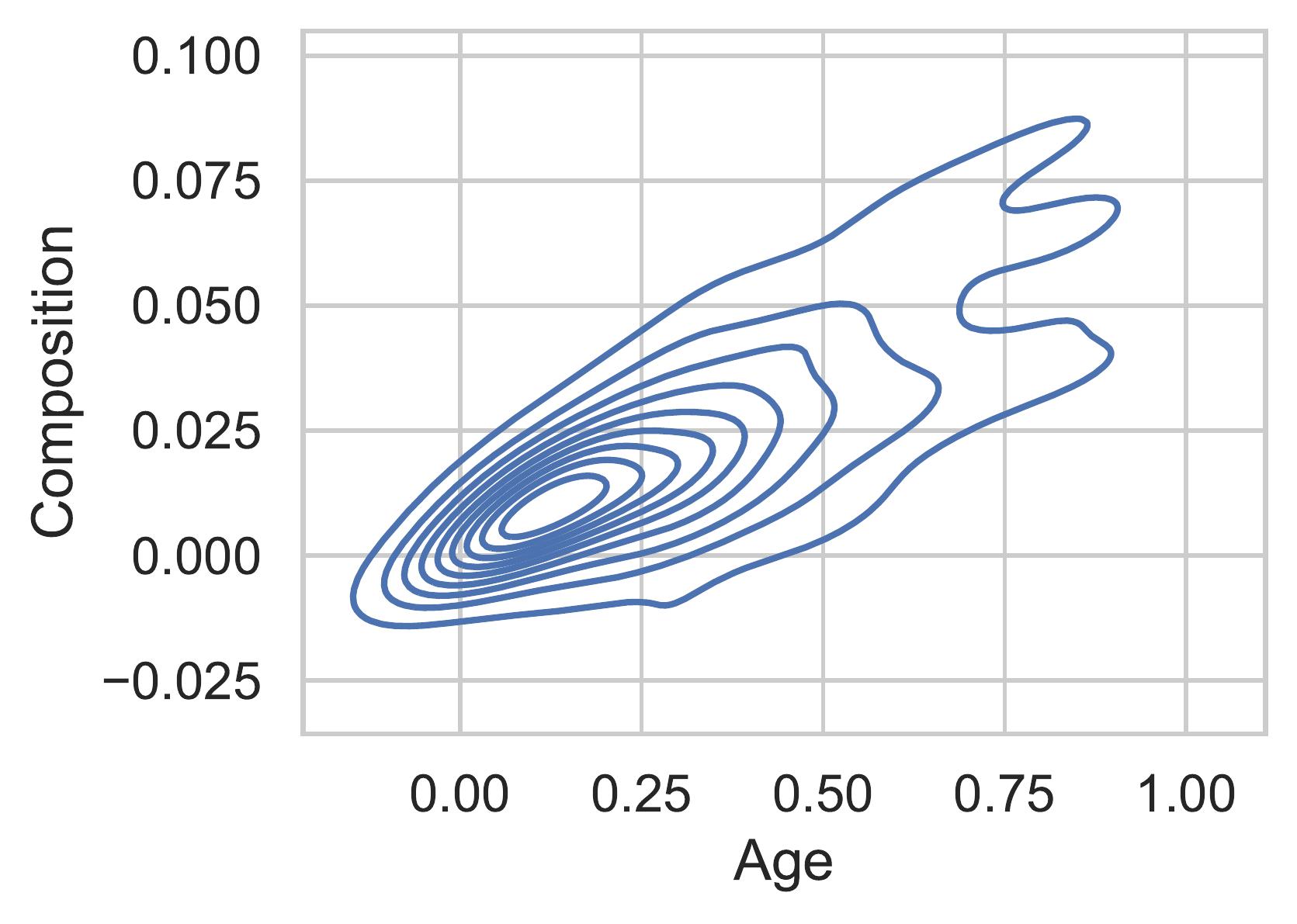}
	\caption{ELU}
	\label{fig:age-good-elu}
	\end{subfigure}
	\begin{subfigure}{0.24\textwidth}
	\center
	\includegraphics[width=\textwidth]{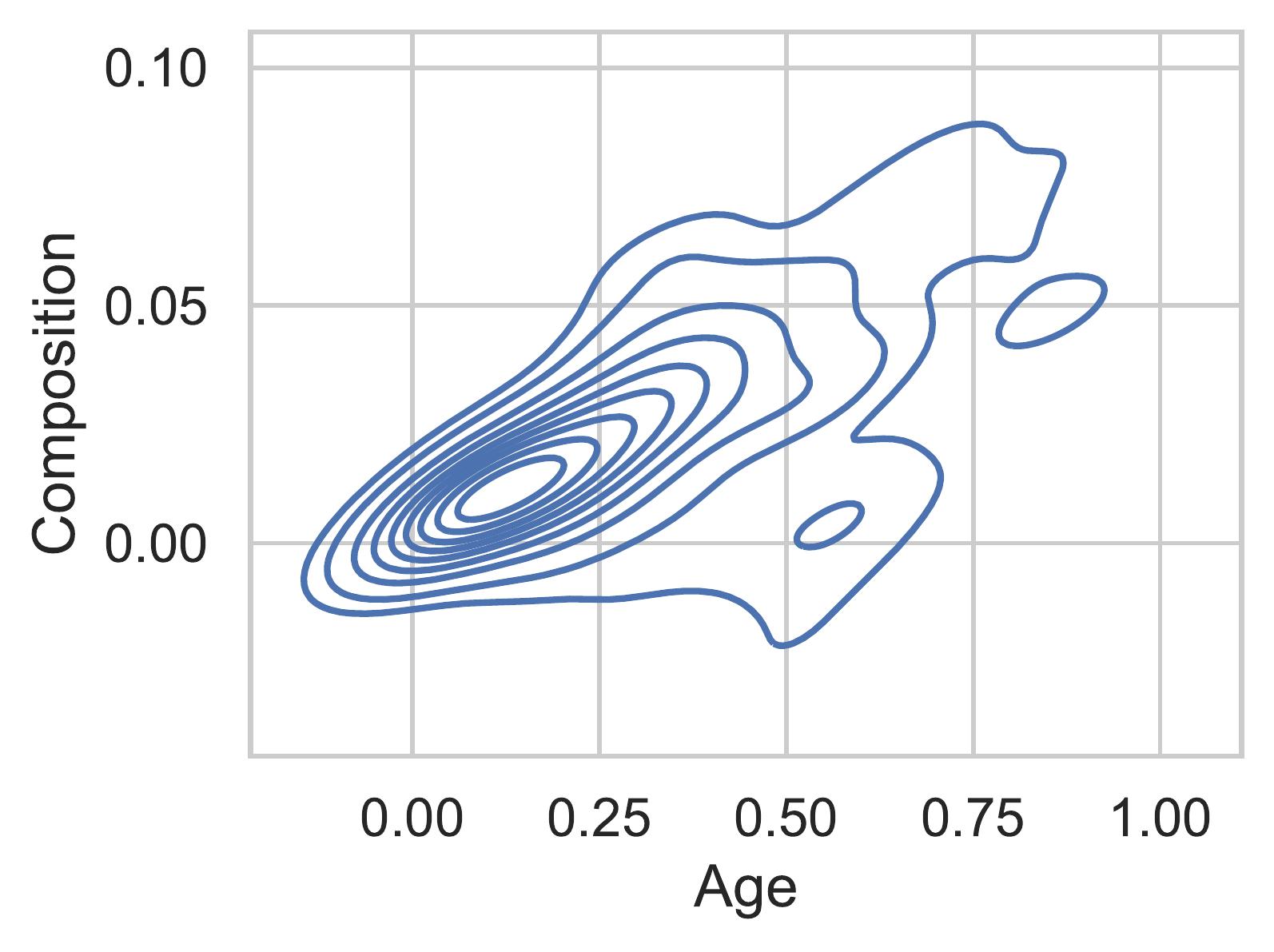}
	\caption{Leaky ReLU}
	\label{fig:age-good-lelu}
	\end{subfigure}
	\begin{subfigure}{0.24\textwidth}
	\center
	\includegraphics[width=\textwidth]{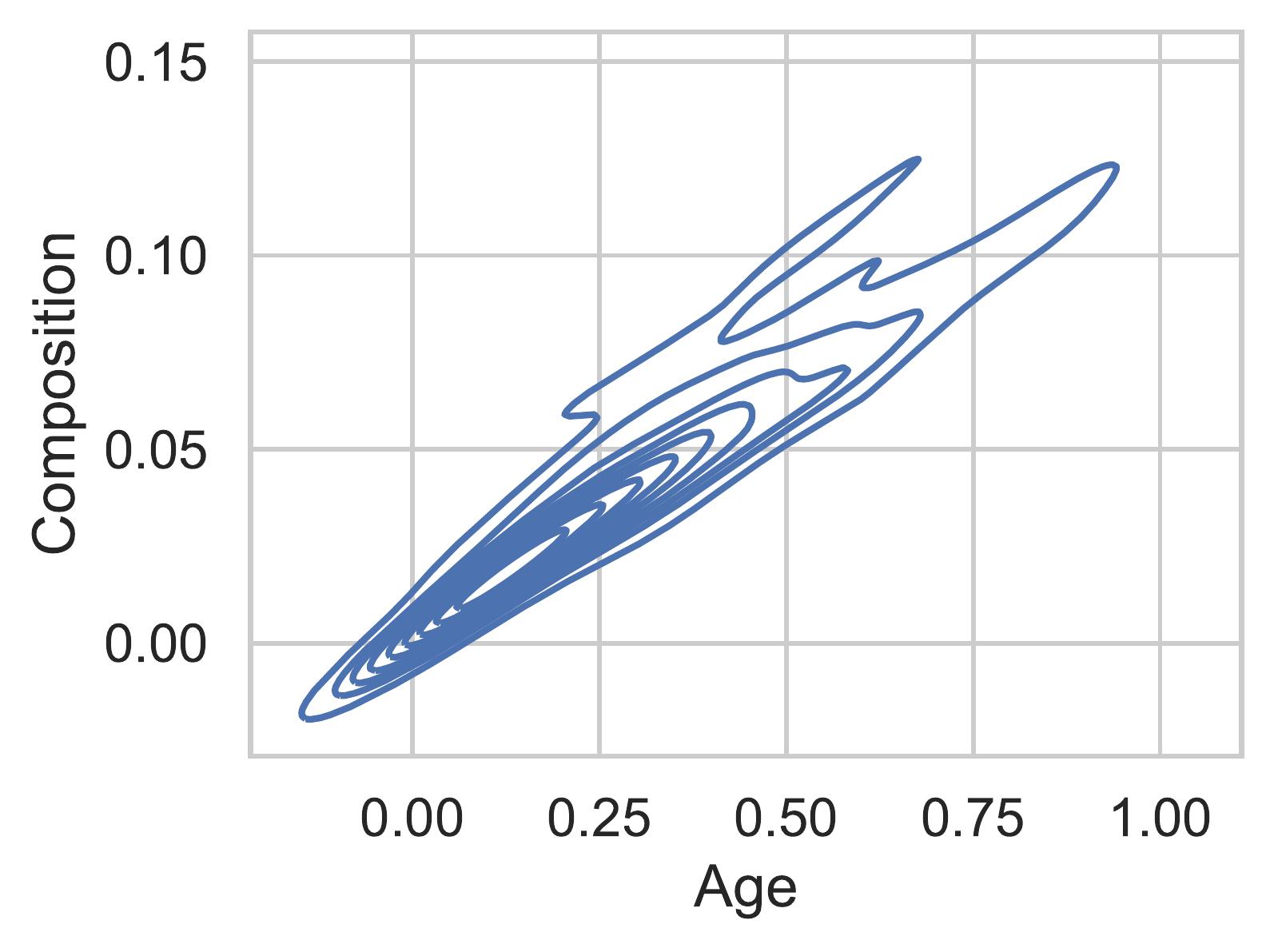}
	\caption{Sigmoid} 
	\label{fig:age-good-sigmoid}
	\end{subfigure}
	\begin{subfigure}{0.24\textwidth}
	\center
	\includegraphics[width=\textwidth]{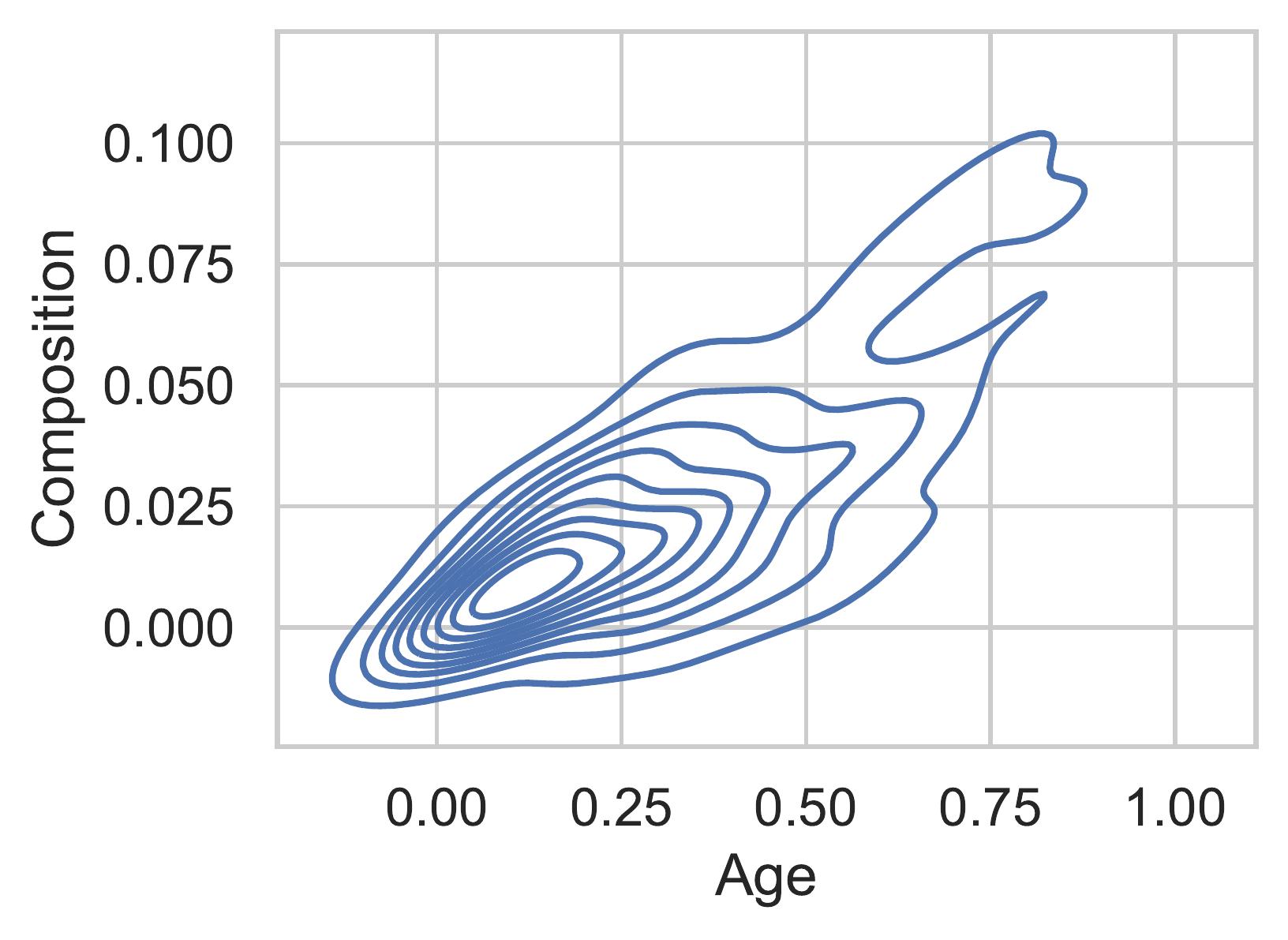}
	\caption{Hyperbolic Tangent}
	\label{fig:age-good-tanh}
	\end{subfigure}
	
	\captionsetup{justification=justified}
	\caption{Composition values for the protected feature \textit{age} contrasted with its values, for all instances predicted as good credit. There seems to be a positive correlation between the values of that feature and the composition values computed by the FCP algorithm, i.e., the older the person, the higher and more positive the influence of \textit{age} in the outcome for good credit.}
\label{fig:age-good}
\end{figure*}

\begin{figure*}[!htbp]
\center
    \begin{subfigure}{0.24\textwidth}
	\center
	\includegraphics[width=\textwidth]{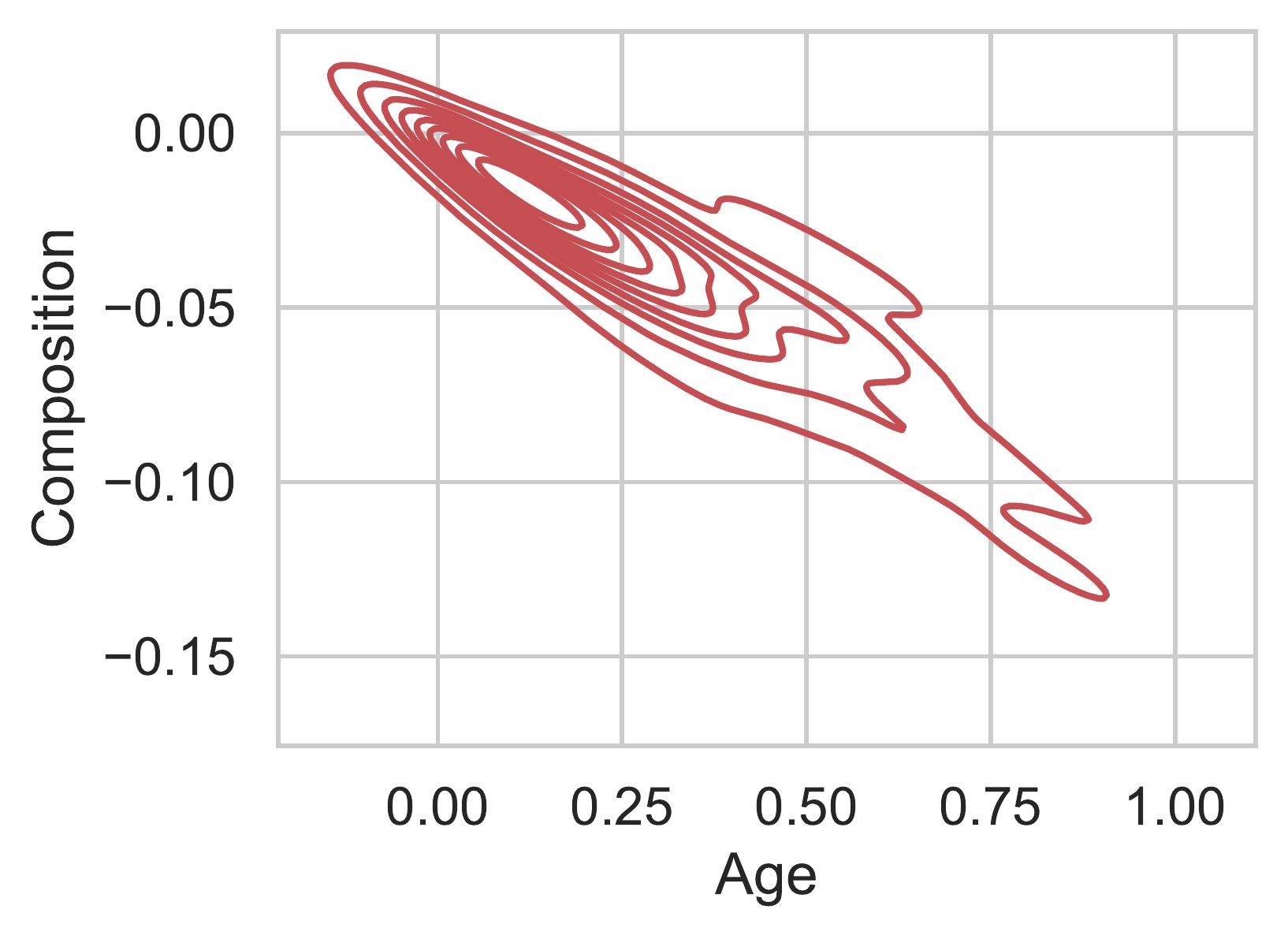}
	\caption{ELU}
	\label{fig:age-bad-elu}
	\end{subfigure}
	\begin{subfigure}{0.24\textwidth}
	\center
	\includegraphics[width=\textwidth]{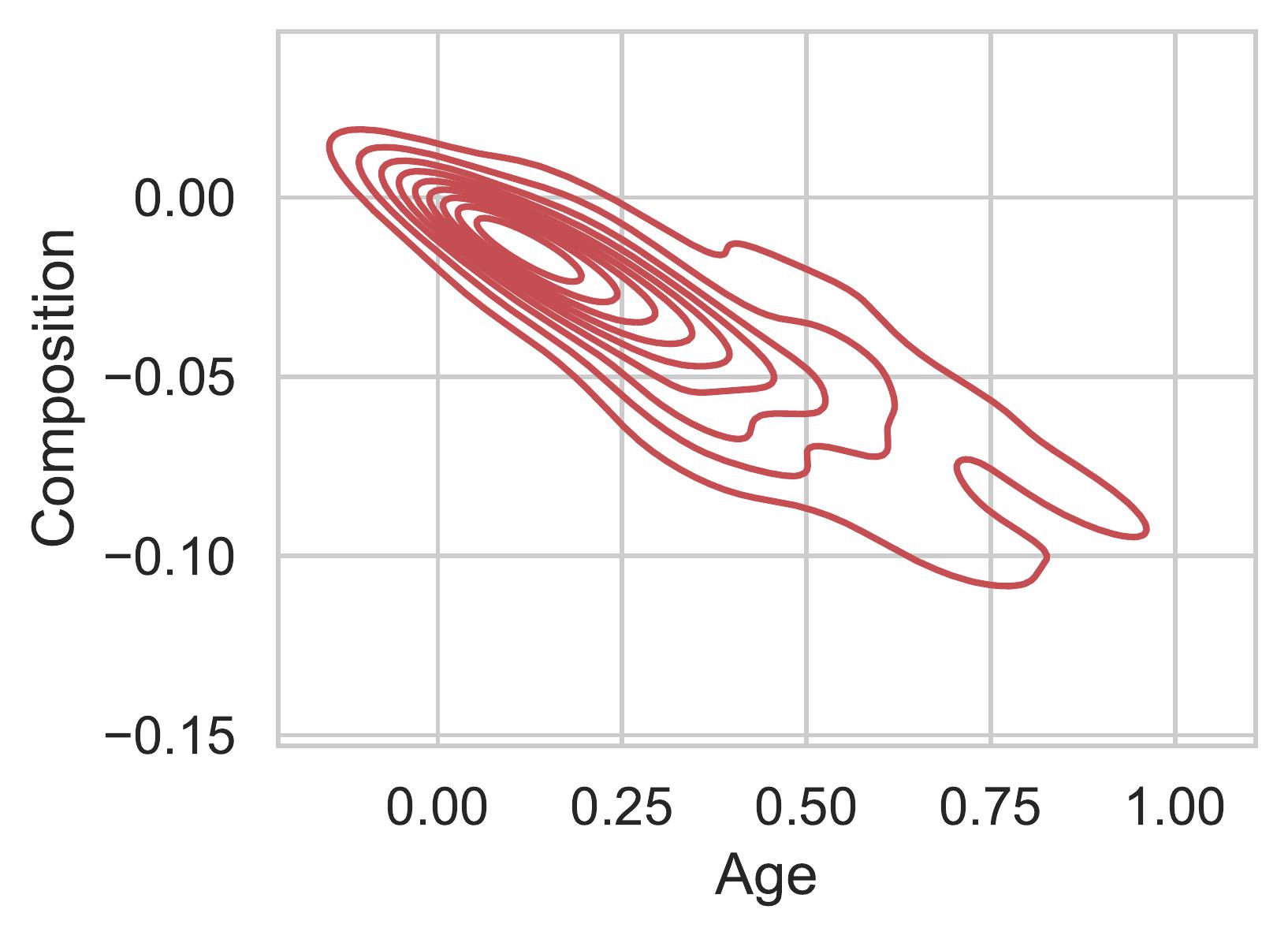}
	\caption{Leaky ReLU}
	\label{fig:age-bad-lelu}
	\end{subfigure}
	\begin{subfigure}{0.24\textwidth}
	\center
	\includegraphics[width=\textwidth]{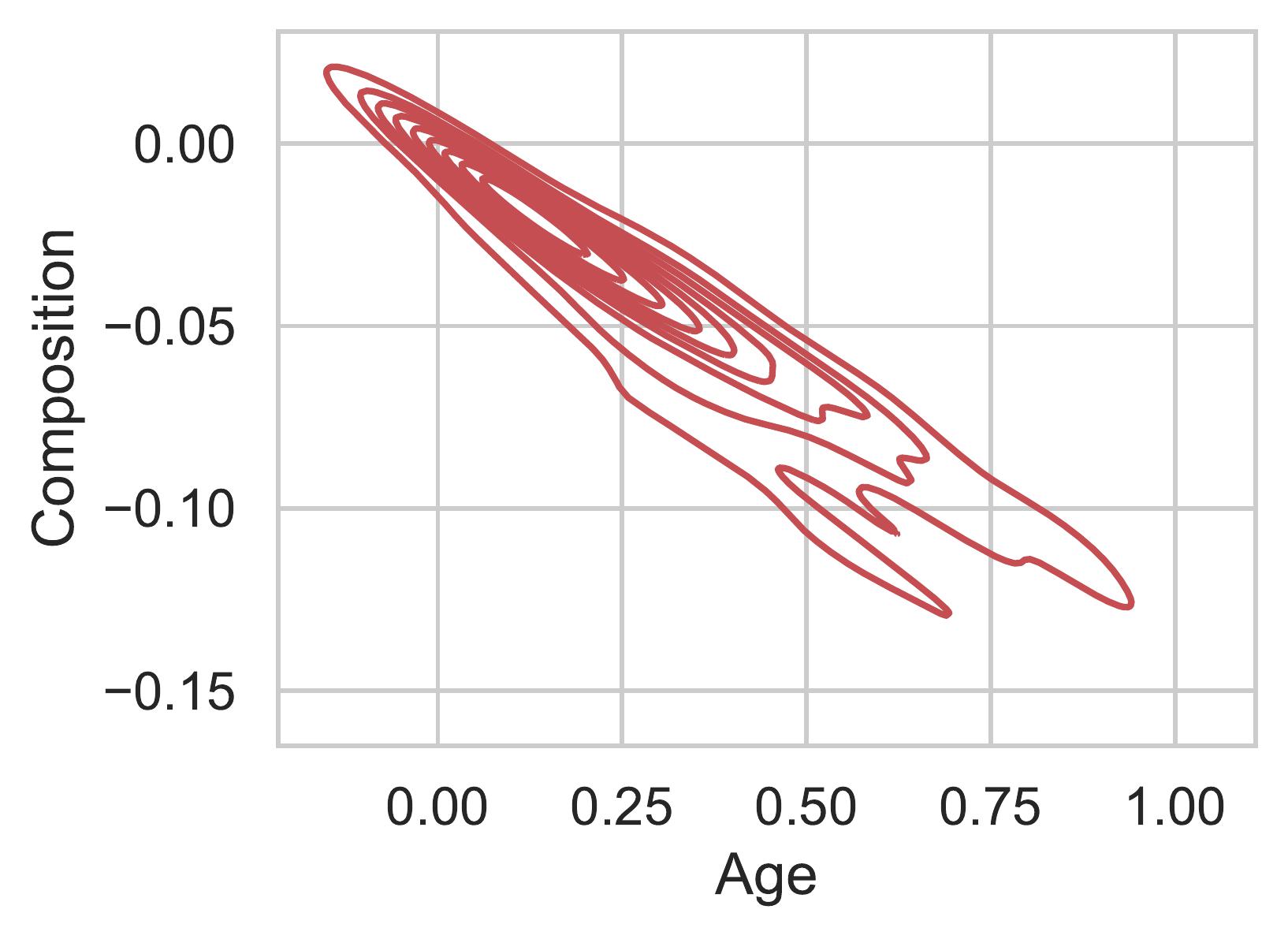}
	\caption{Sigmoid}
	\label{fig:age-bad-sigmoid}
	\end{subfigure}
	\begin{subfigure}{0.24\textwidth}
	\center
	\includegraphics[width=\textwidth]{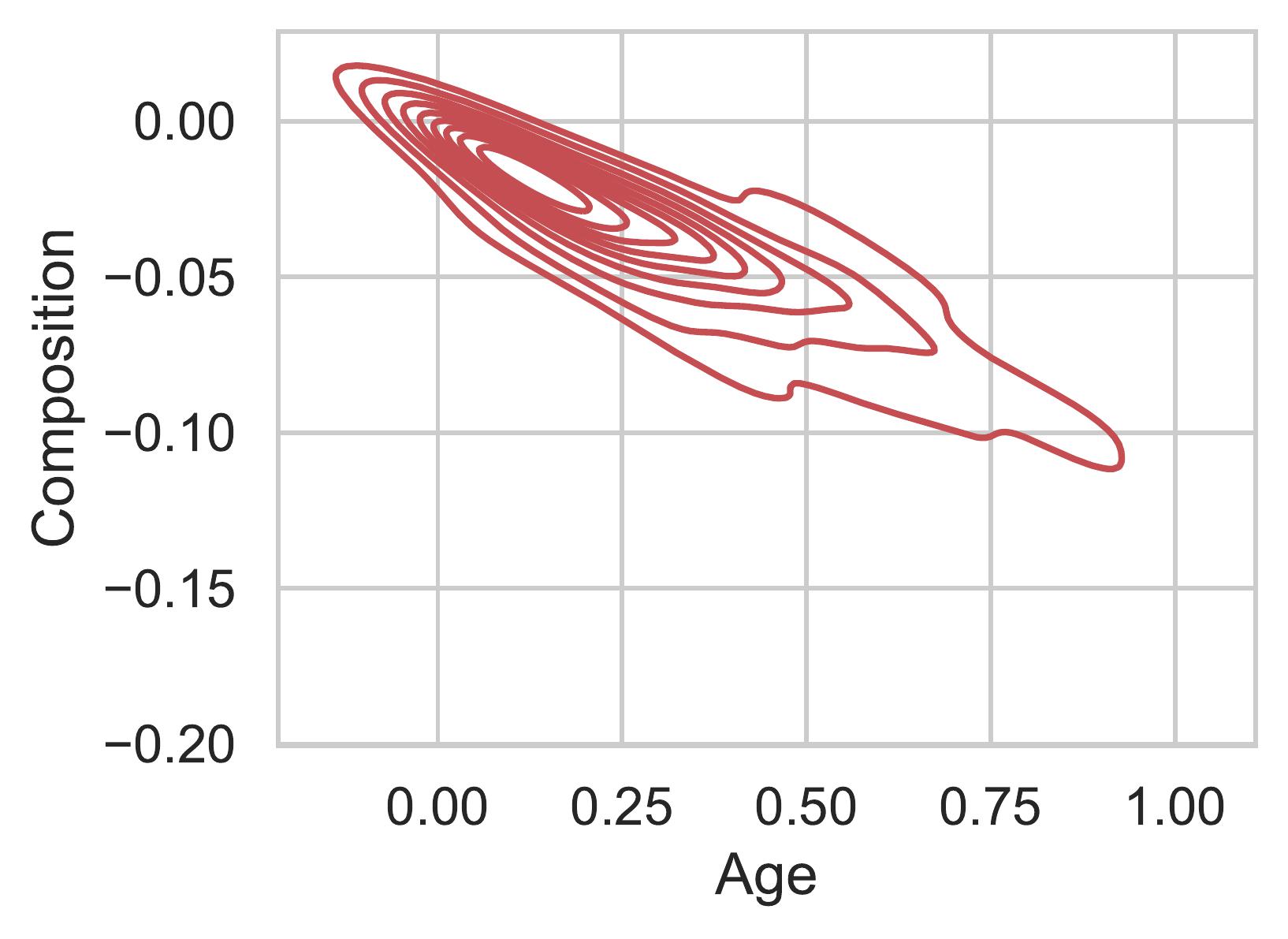}
	\caption{Hyperbolic Tangent}
	\label{fig:age-bad-tanh}
	\end{subfigure}	
	
	\captionsetup{justification=justified}
	\caption{Composition values for the protected feature \textit{age} contrasted with its values, for all instances predicted as bad credit. There seems to be a negative correlation between the values of that feature and the composition values computed by the FCP algorithm, i.e., the older the person, the more negative the influence of \textit{age} in the outcome for bad credit.}
\label{fig:age-bad}
\end{figure*}

The first step in the analysis is inspecting the composition values attached to the protected feature \textit{age} for the cases of good credit (see Figure \ref{fig:age-good}) and bad credit (see Figure \ref{fig:age-bad}) using different transfer functions. These density plots contrast the composition values computed with our algorithm and the normalized values of the protected feature. There seem to be strong linear correlation patterns between the feature values and the composition values even when the FCP algorithm operates with the activation values of neurons, which come out from a non-linear transfer function.

Table \ref{tab:correlation} shows the Pearson's correlation between the feature under analysis and the corresponding composition values for both good and bad credits. Overall, the results show a positive correlation between the composition values and the feature values in the case of good credit. This means that the model favors older applicants with good credits, and the composition values capture such a pattern. Similarly, the results indicate a negative correlation between the composition values and the feature values in the case of bad credit. This behavior matches with the ground truth suggesting a tendency against granting bad credit to older people.

\begin{table}[!htbp]
\caption{Pearson's correlation between the protected feature age and the corresponding composition values for both good and bad credits.}
\centering
\label{tab:correlation}
\begin{tabular}{|c|c|c|c|c|}
\hline
            & ELU    & Leaky ReLU & Sigmoid & Hyperbolic \\ \hline
Good credit & 0.712  & 0.659      & 0.951   & 0.753      \\ \hline
Bad credit  & -0.911 & -0.832     & -0.951  & -0.879     \\ \hline
\end{tabular}
\end{table}

The results in Figures \ref{fig:age-good} and \ref{fig:age-bad} also indicate that the sigmoid function reports the largest correlation between the feature values and the composition values associated with the protected feature age.

\begin{figure*}[!htbp]
\center
    \begin{subfigure}{0.24\textwidth}
	\center
	\includegraphics[width=\textwidth]{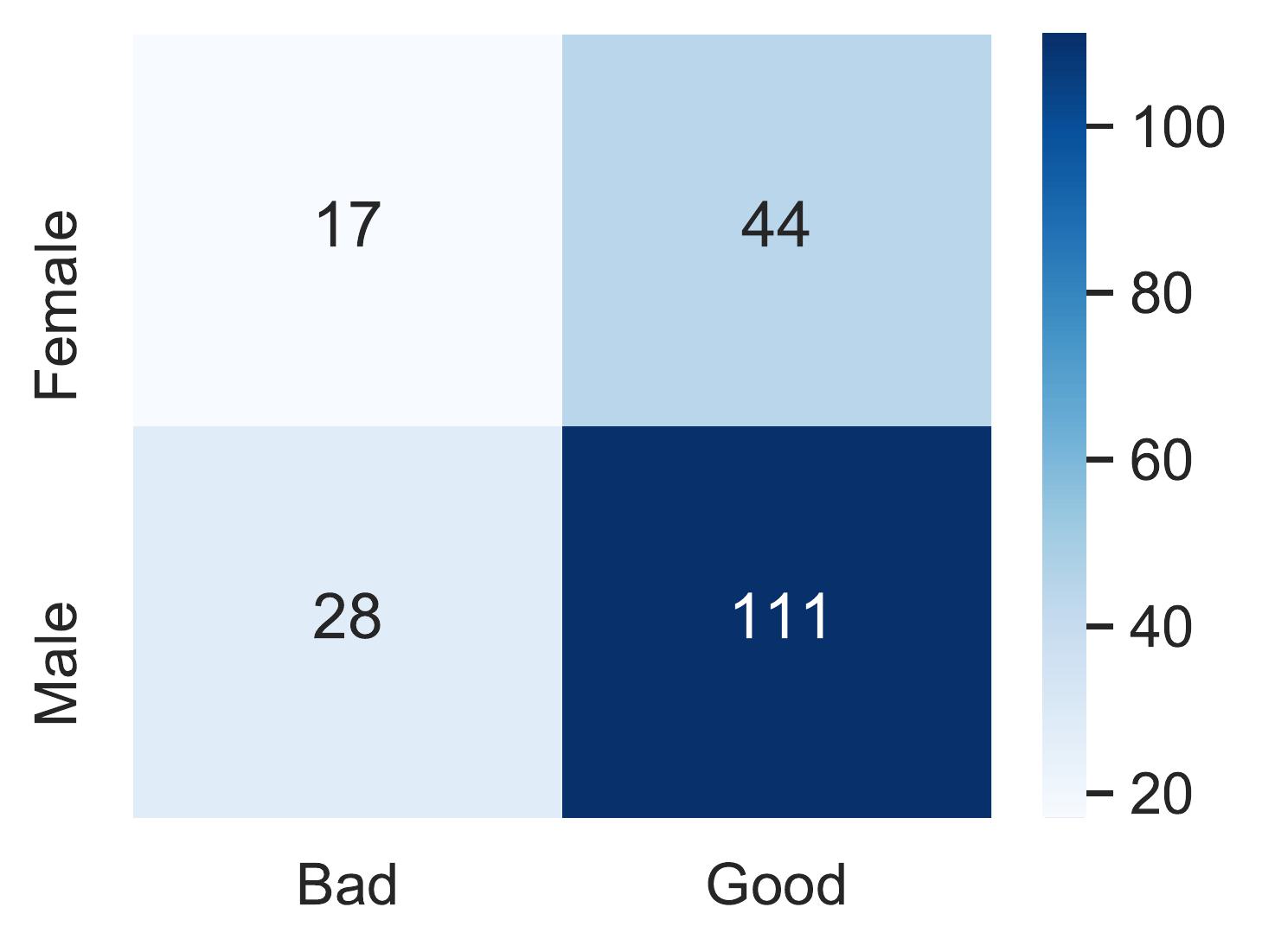}
	\caption{ELU, $\kappa=0.49$}
	\label{fig:predictions-elu}
	\end{subfigure}
	\begin{subfigure}{0.24\textwidth}
	\center
	\includegraphics[width=\textwidth]{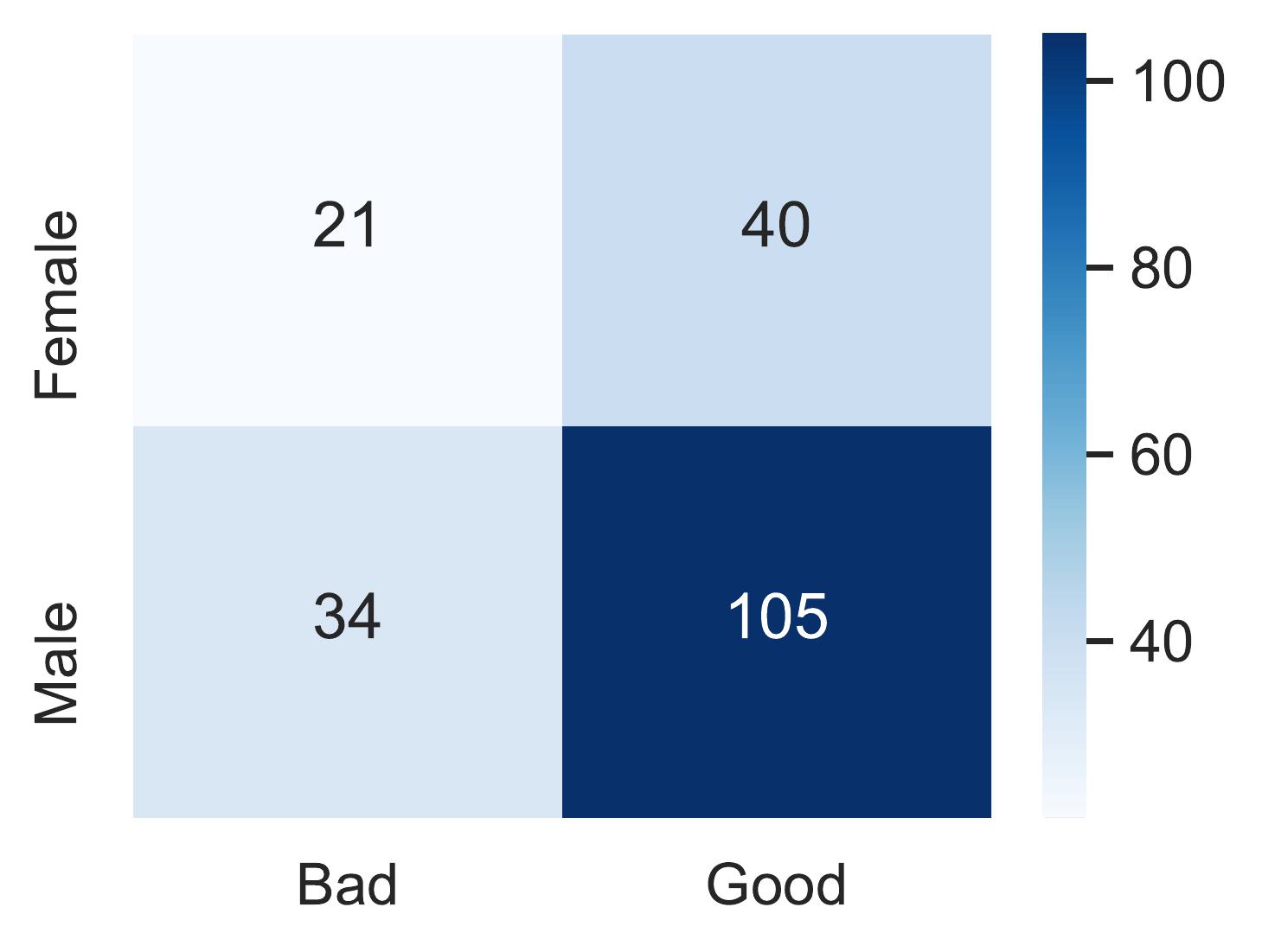}
	\caption{Leaky ReLU, $\kappa=0.40$}
	\label{fig:predictions-lelu}
	\end{subfigure}
	\begin{subfigure}{0.24\textwidth}
	\center
	\includegraphics[width=\textwidth]{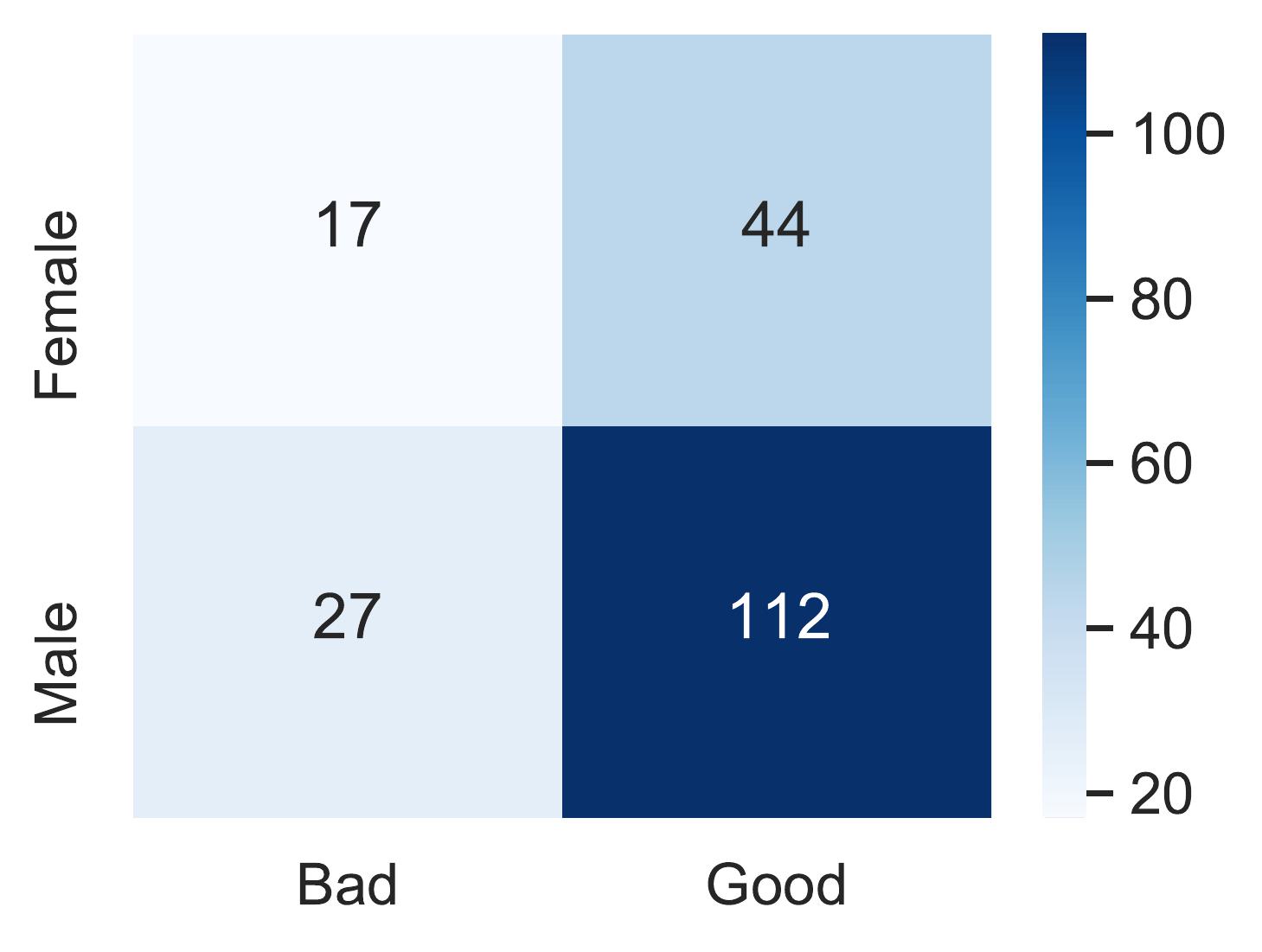}
	\caption{Sigmoid, $\kappa=0.09$}
	\label{fig:predictions-sigmoid}
	\end{subfigure}
	\begin{subfigure}{0.24\textwidth}
	\center
	\includegraphics[width=\textwidth]{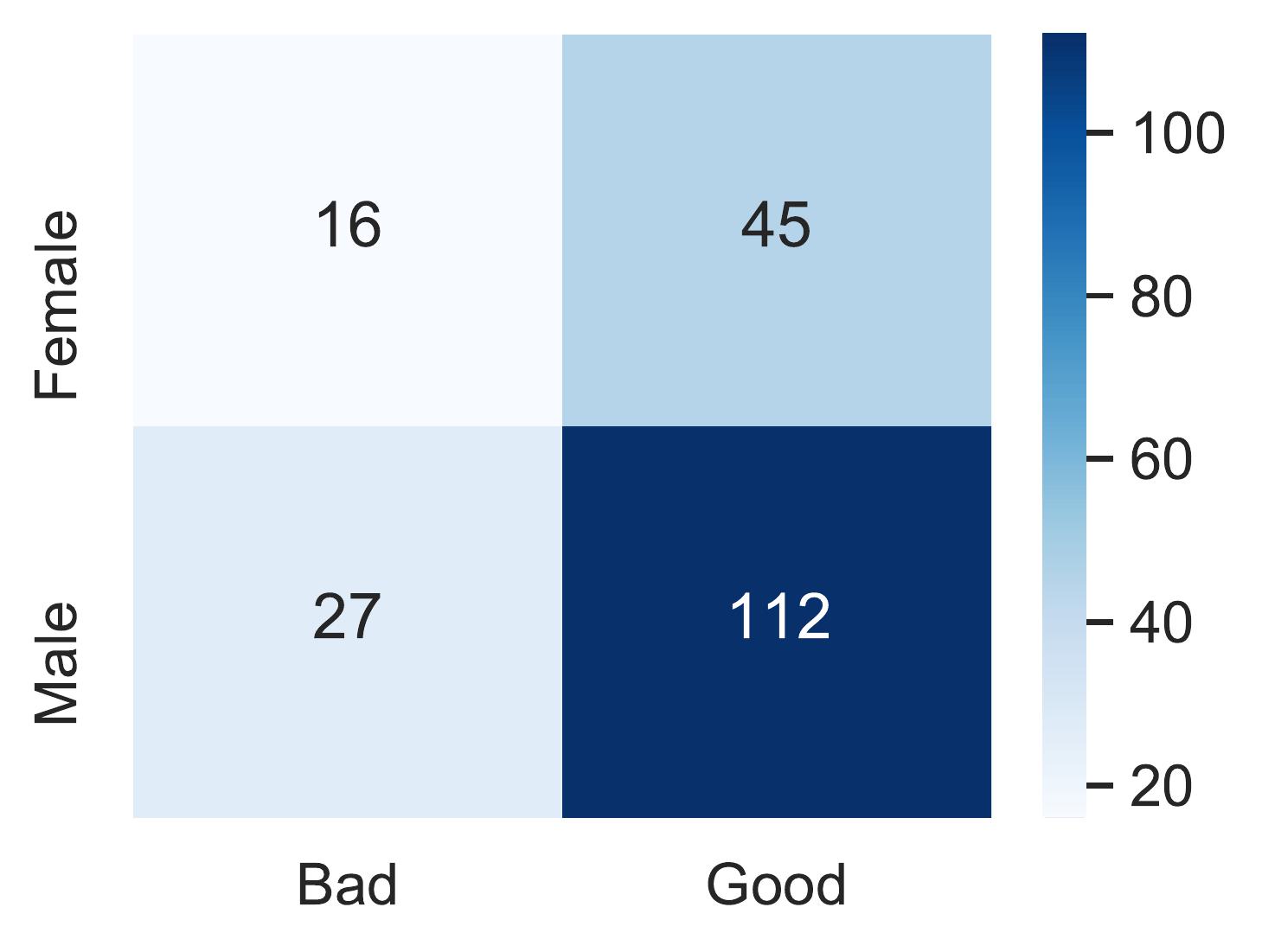}
	\caption{Hyperbolic Tan., $\kappa=0.38$}
	\label{fig:predictions-tanh}
	\end{subfigure}	
	
	\captionsetup{justification=justified}
	\caption{Distribution of good and bad credits for both female and male applicants as computed by the neural network on the test data. It shall be noticed that male applicants are more likely to receive good credit when considering only the feature gender. These patterns will be compared with the decisions derived from composition values associated with the feature gender in the output layer.}
\label{fig:predictions}
\end{figure*}

\begin{figure*}[!htbp]
\center
    \begin{subfigure}{0.24\textwidth}
	\center
	\includegraphics[width=\textwidth]{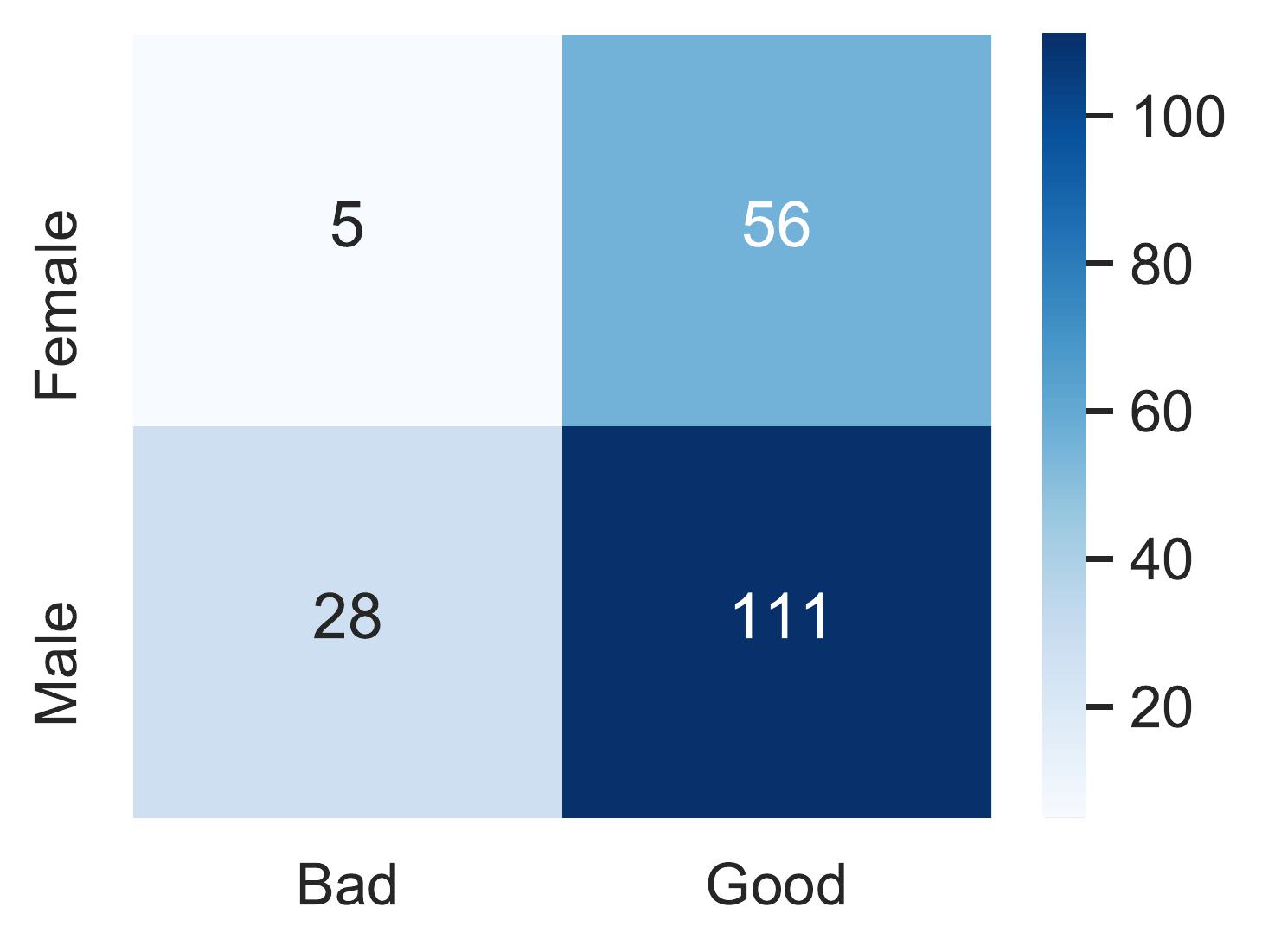}
	\caption{ELU, $\kappa=0.49$}
	\label{fig:compositions-elu}
	\end{subfigure}
	\begin{subfigure}{0.24\textwidth}
	\center
	\includegraphics[width=\textwidth]{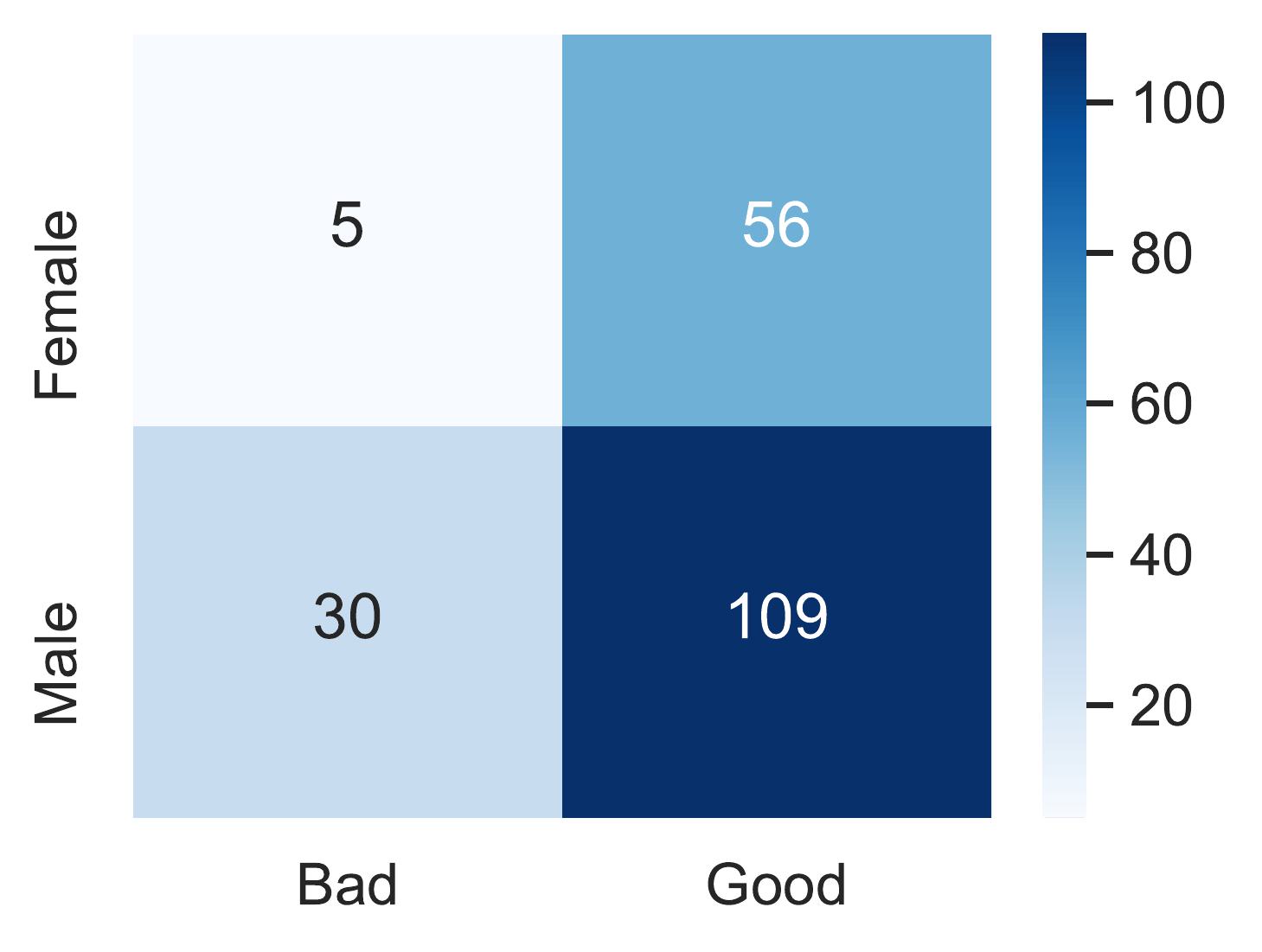}
	\caption{Leaky ReLU, $\kappa=0.40$}
	\label{fig:compositions-relu}
	\end{subfigure}
	\begin{subfigure}{0.24\textwidth}
	\center
	\includegraphics[width=\textwidth]{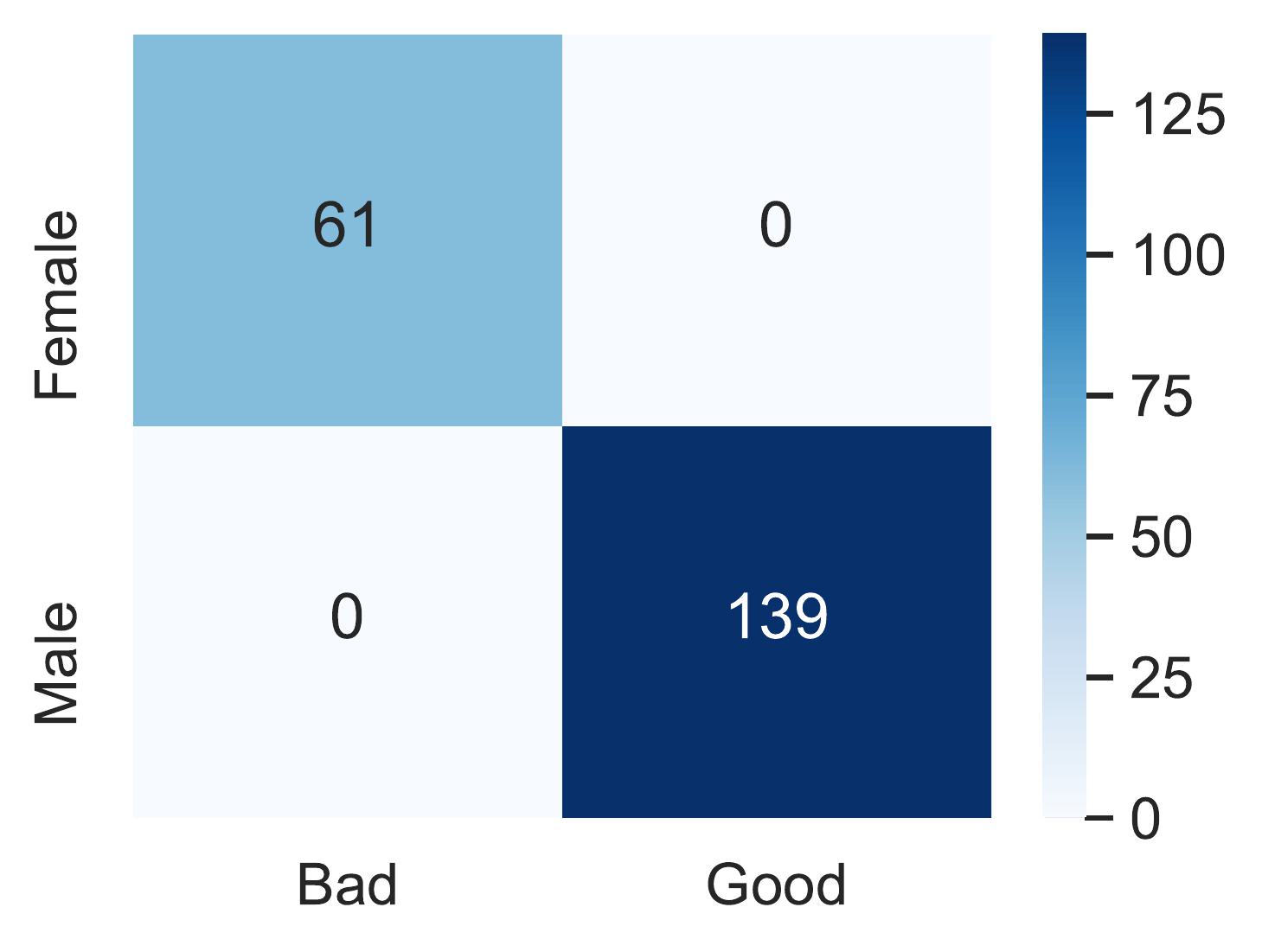}
	\caption{Sigmoid, $\kappa=0.09$}
	\label{fig:compositions-sigmoid}
	\end{subfigure}
	\begin{subfigure}{0.24\textwidth}
	\center
	\includegraphics[width=\textwidth]{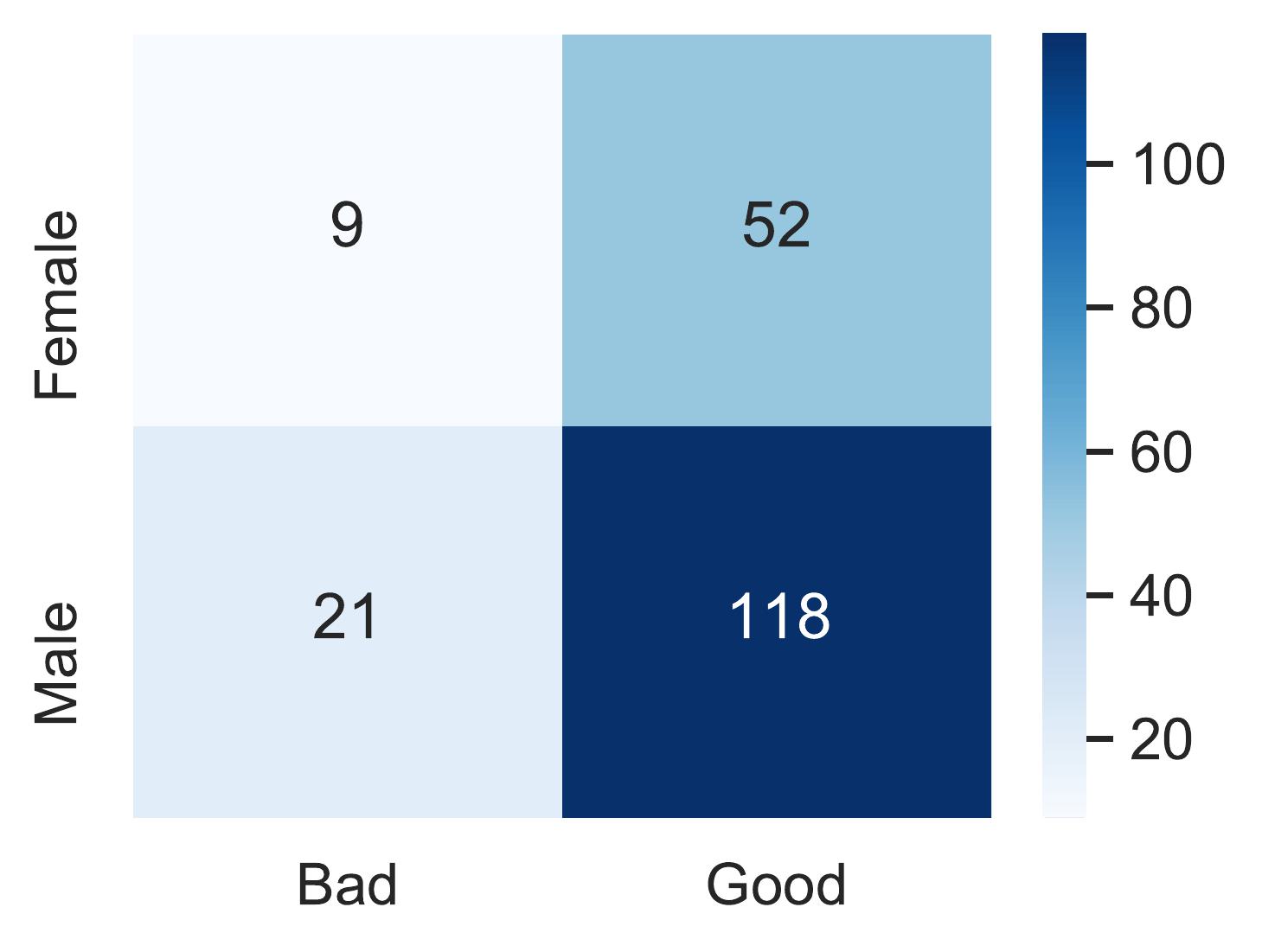}
	\caption{Hyperbolic Tan., $\kappa=0.38$}
	\label{fig:compositions-tanh}
	\end{subfigure}	
	
	\captionsetup{justification=justified}
	\caption{Distribution of good and bad credits for both female and male applicants computed from the composition vectors attached to the test instances. Similar to the patterns obtained with the neural network, male applicants are more likely to receive good credit when considering only the feature gender.}
\label{fig:compositions}
\end{figure*}

The analysis continues by inspecting the composition values attached to the protected feature \textit{gender} in the case of good and bad credits for different transfer functions. Figure \ref{fig:predictions} shows the distribution of good and bad credits for both female and male applicants as computed by the neural network. These values will be used for reference as they are approximations of the ground truth obtained with different transfer functions. Figure \ref{fig:compositions} displays the distribution of good and bad credits for both gender categories as computed from the composition values in the last layer.

In this experiment, Cohen's kappa coefficient ($\kappa$) \cite{kappa1960} is adopted to measure the agreement in the predictions computed from the neural network and the ones computed from the composition values. Like other correlation statistics, kappa values range from -1 to +1 such that positive values give a measure of agreement. The kappa values when using ELU, Leaky ReLU, sigmoid, and hyperbolic tangent as the activation functions are 0.49, 0.40, 0.09, and 0.38, respectively. These values suggest that composition values can be used to capture the bias patterns encoded in the data. It shall be mentioned that a perfect agreement was not expected since the neural network makes the decisions using all features, not only the protected one.

The results concerning the sigmoid function require further discussion since its corresponding kappa value is close to zero. When using this function, the predictions computed from the composition values align perfectly with the gender categories (i.e., males will receive a good credit while females will receive a bad one). As mentioned above, the low agreement between the predictions computed by the network and those computed from compositions suggests that the network considers other features. Although these features provide some balance in the decisions, the predictions continue to be biased toward granting good credits to male applicants.

\subsection{Comparing against LRP and SHAP feature attributions}
\label{sec:simulations:comparison}

Although our algorithm is not intended for feature importance analysis, the feature rankings extracted from the composition values can be used as a proxy for its validation. This section qualitatively compares the compositions extracted by the proposed FCP algorithm with the relevance values computed by the LRP algorithm \cite{Montavon2019} and the feature attribution extracted by the SHAP method \cite{NIPS2017_7062}. It is worth noticing that the numerical quantities of feature attributions computed by these methods are not directly comparable.

\begin{figure}[!htbp]
\center
    \begin{subfigure}{\columnwidth}
	\center
	\includegraphics[width=\columnwidth]{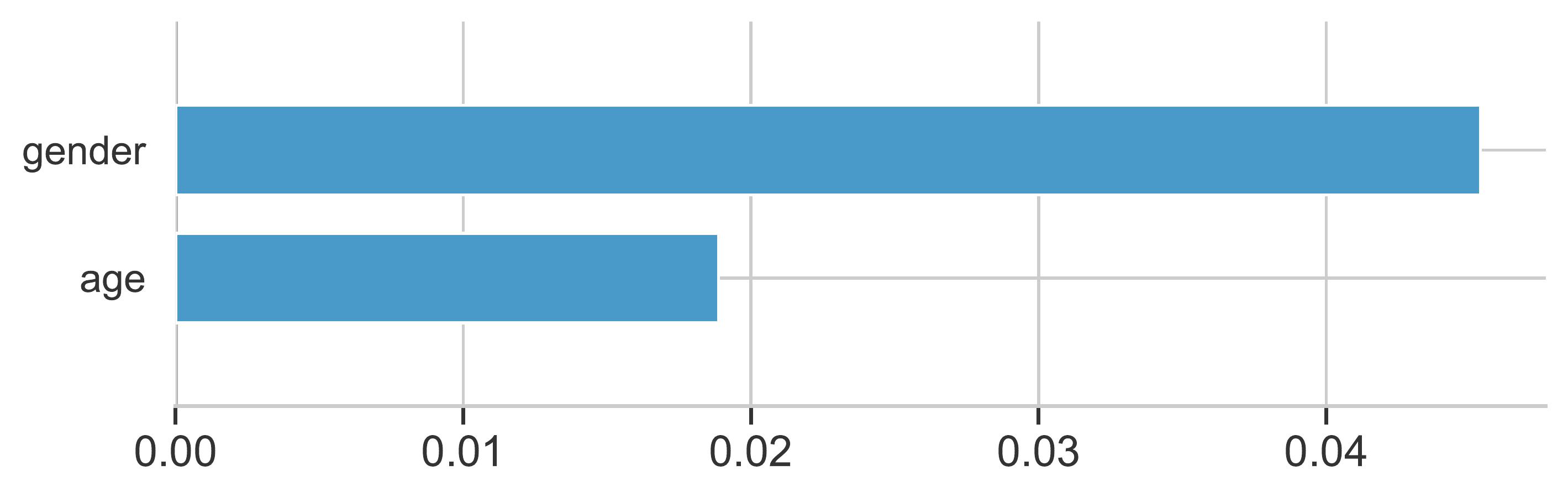}
	\caption{FCP}
	\label{fig:comp-fcp}
	\end{subfigure}
	\begin{subfigure}{\columnwidth}
	\center
	\includegraphics[width=\columnwidth]{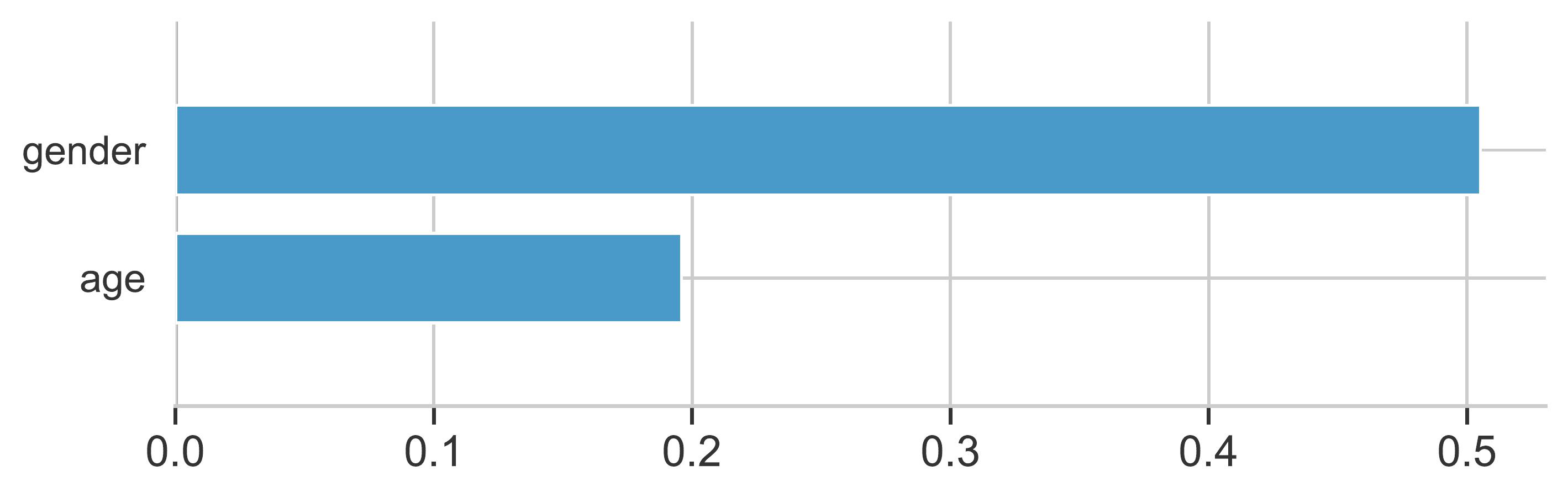}
	\caption{LRP}
	\label{fig:comp-LRP}
	\end{subfigure}
	\begin{subfigure}{\columnwidth}
	\center
	\includegraphics[width=\columnwidth]{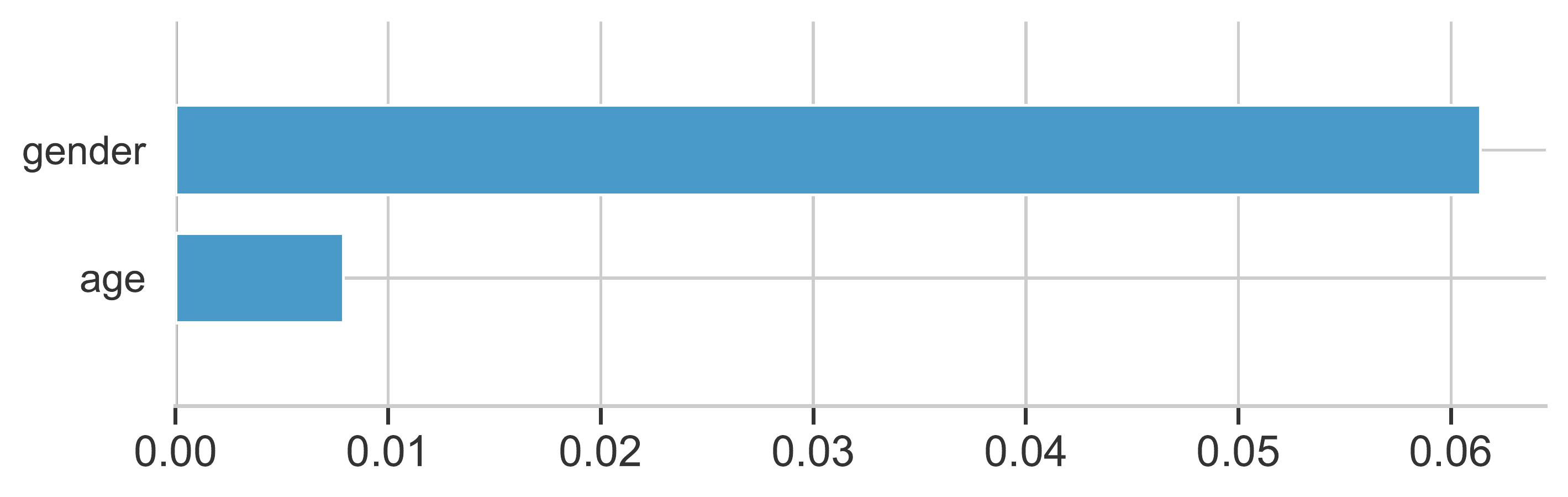}
	\caption{SHAP}
	\label{fig:comp-shap}
	\end{subfigure}
	
	\captionsetup{justification=justified}
	\caption{Mean feature attribution values associated with \textit{gender} and \textit{age}, computed by each method, over all instances.}
\label{fig:comparison}
\end{figure}

As mentioned before, the LRP algorithm \cite{Montavon2019} is a local, model-specific approach to compute feature attribution for neural networks in the form of backpropagated relevance values. SHAP is a local and model-agnostic approach for computing feature attribution based on the contribution that a feature brings to all subsets of features when making the prediction.  Since these explanation methods are local, the global estimates of the feature attribution values will be obtained by averaging the absolute values over all instances. In this experiment, the underlying neural network model maintains the same parameter configuration as the previous experiments while using the ReLU activation function to make the results comparable with the LRP algorithm. The LRP implementation uses the $\epsilon$ rule with $\epsilon=1.0E-9$. Figure \ref{fig:comparison} shows the feature attribution values computed by FCP, LRP, and SHAP for the protected features \textit{gender} and \textit{age}.

Figure \ref{fig:comparison} shows that all methods assign more attribution to the feature \textit{gender} compared to \textit{age}, which means that the underlying neural network relies more on \textit{gender} than \textit{age} to make the predictions. In addition, these results are in agreement with the fuzzy-rough set measure for explicit and implicit bias proposed in \cite{NAPOLES202229} and the results of the implicit bias simulations using Fuzzy Cognitive Maps proposed in \cite{NAPOLES202233}. Therefore, these results confirm the reliability of the compositions computed by our FCP algorithm. At the same time, we want to highlight that, unlike LRP or SHAP, FCP feature attributions are obtained for each unit in the neural network, providing more information when formulating explanations.

\subsection{Extended experiments with benchmark data}
\label{sec:simulations:benchmark}

This section conducts extended simulations using benchmark datasets in the context of feature importance. The ranking of the features based on the FCP compositions is used to perform a validation inspired by the pixel-flipping experiment \cite{samek2016evaluating}. In our version, instead of pixels, we gradually remove the information of features ranked by their FCP composition value and measure how quickly the prediction score decreases in consequence. This experiment uses eight real-world structured classification problems taken from the UCI Machine Learning \cite{dua2017uci} and OpenML \cite{vanschoren2014openml} repositories. These problems are fairly diverse, with the number of decision classes ranging from 2 to 1000, the number of features ranging from 11 to 202, and the number of instances ranging from 194 to 200,000.

The experimental methodology is structured as follows. Firstly, the feature importance is computed for each instance, using the composition vector associated with the output neuron determining the decision class. Secondly, all feature importance scores are averaged to obtain global scores and sort them in descending order. Thirdly, noise is added to each problem feature (following the feature importance ranking) by replacing the feature values with their mean, and the performance degradation is measured. In this way, features are progressively modified to measure the accumulated induced error. This experiment is repeated 20 times to draw more consistent conclusions.

\begin{figure*}[!htbp]
\center
    \begin{subfigure}{0.24\textwidth}
	\center
	\includegraphics[width=\textwidth]{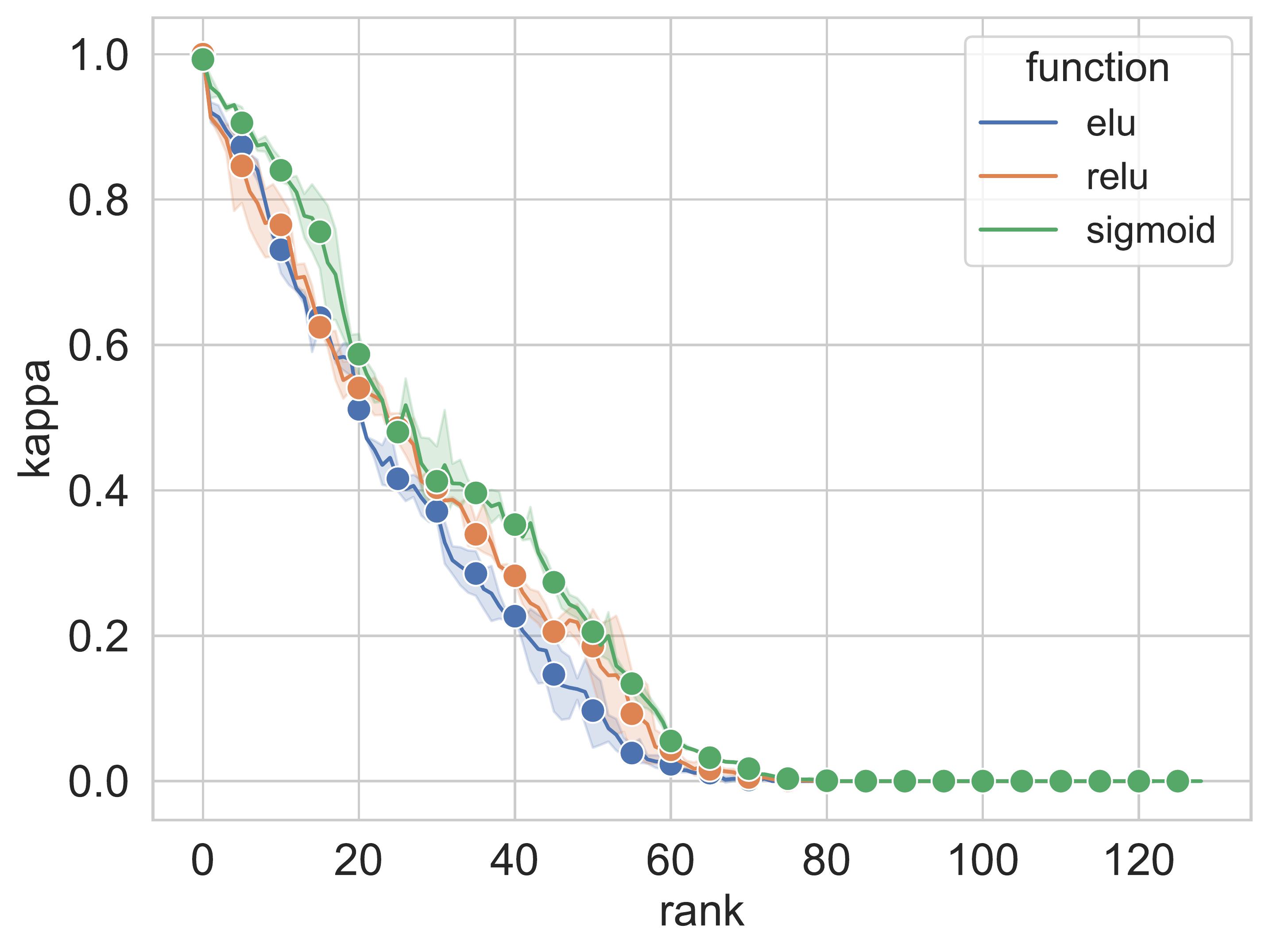}
	\caption{Aloi}
	\end{subfigure}
    \begin{subfigure}{0.24\textwidth}
	\center
	\includegraphics[width=\textwidth]{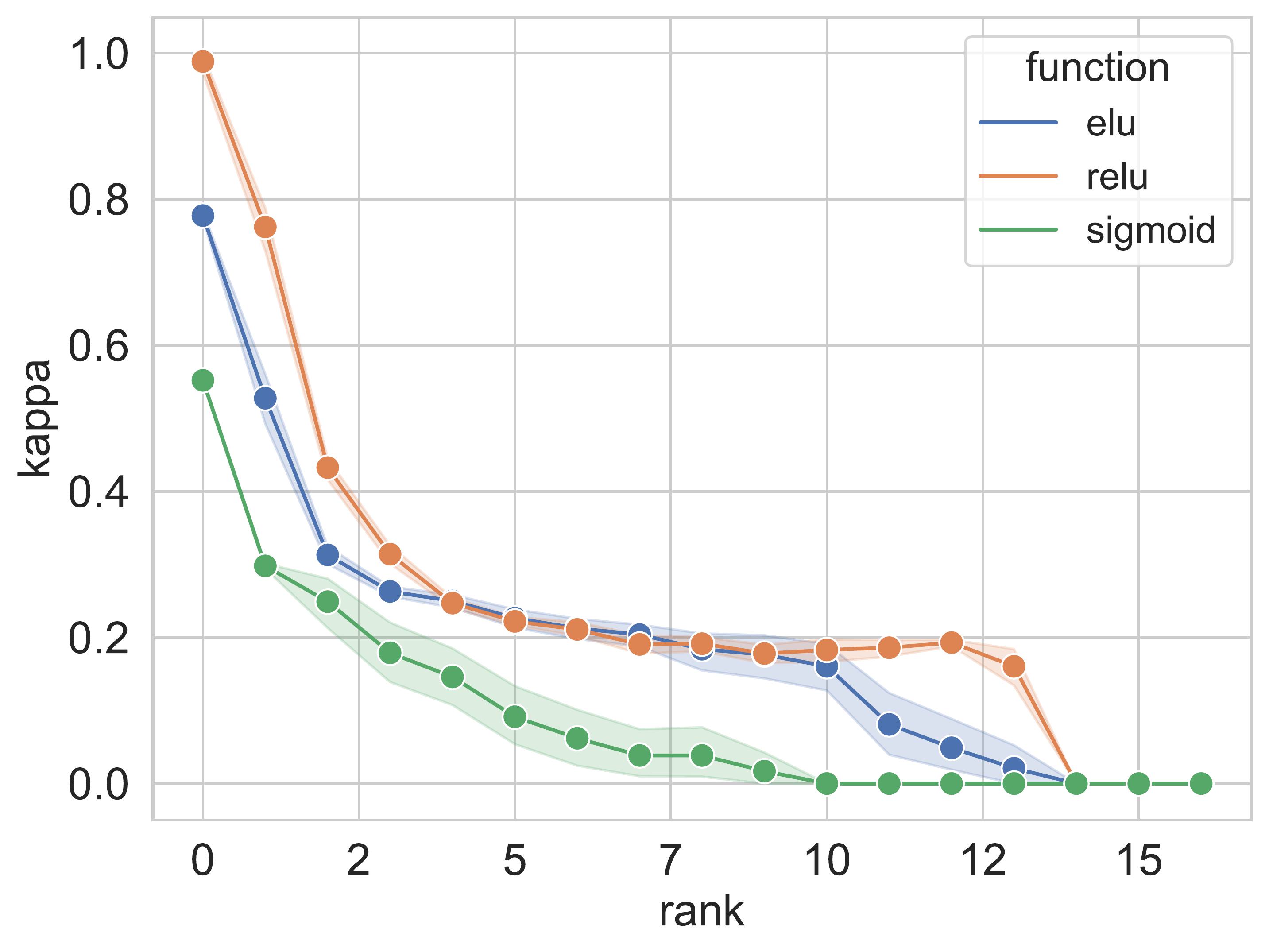}
	\caption{Bank}
	\end{subfigure}
	\begin{subfigure}{0.24\textwidth}
	\center
	\includegraphics[width=\textwidth]{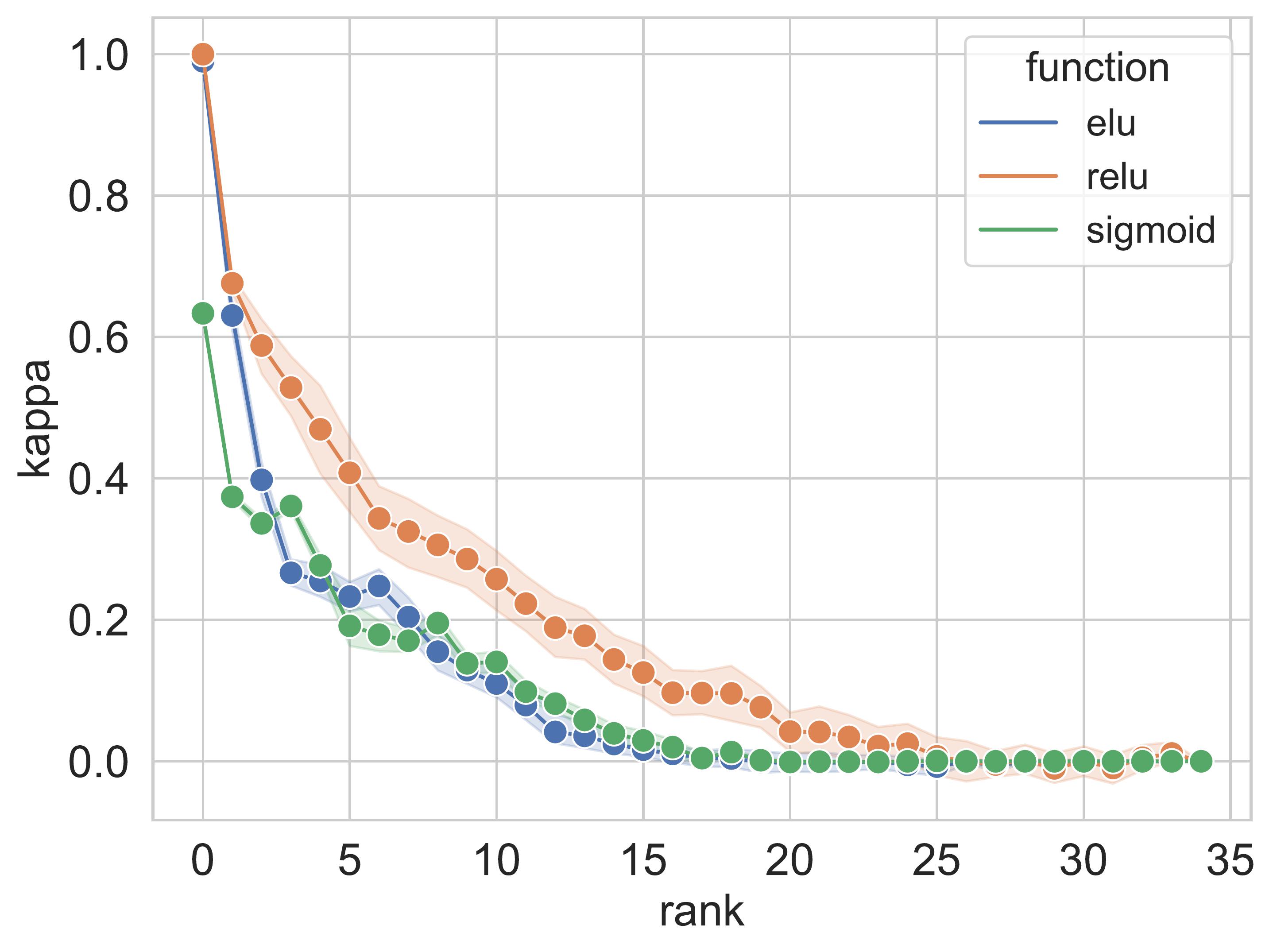}
	\caption{Breast Cancer Wisconsin}
	\end{subfigure}
	\begin{subfigure}{0.24\textwidth}
	\center
	\includegraphics[width=\textwidth]{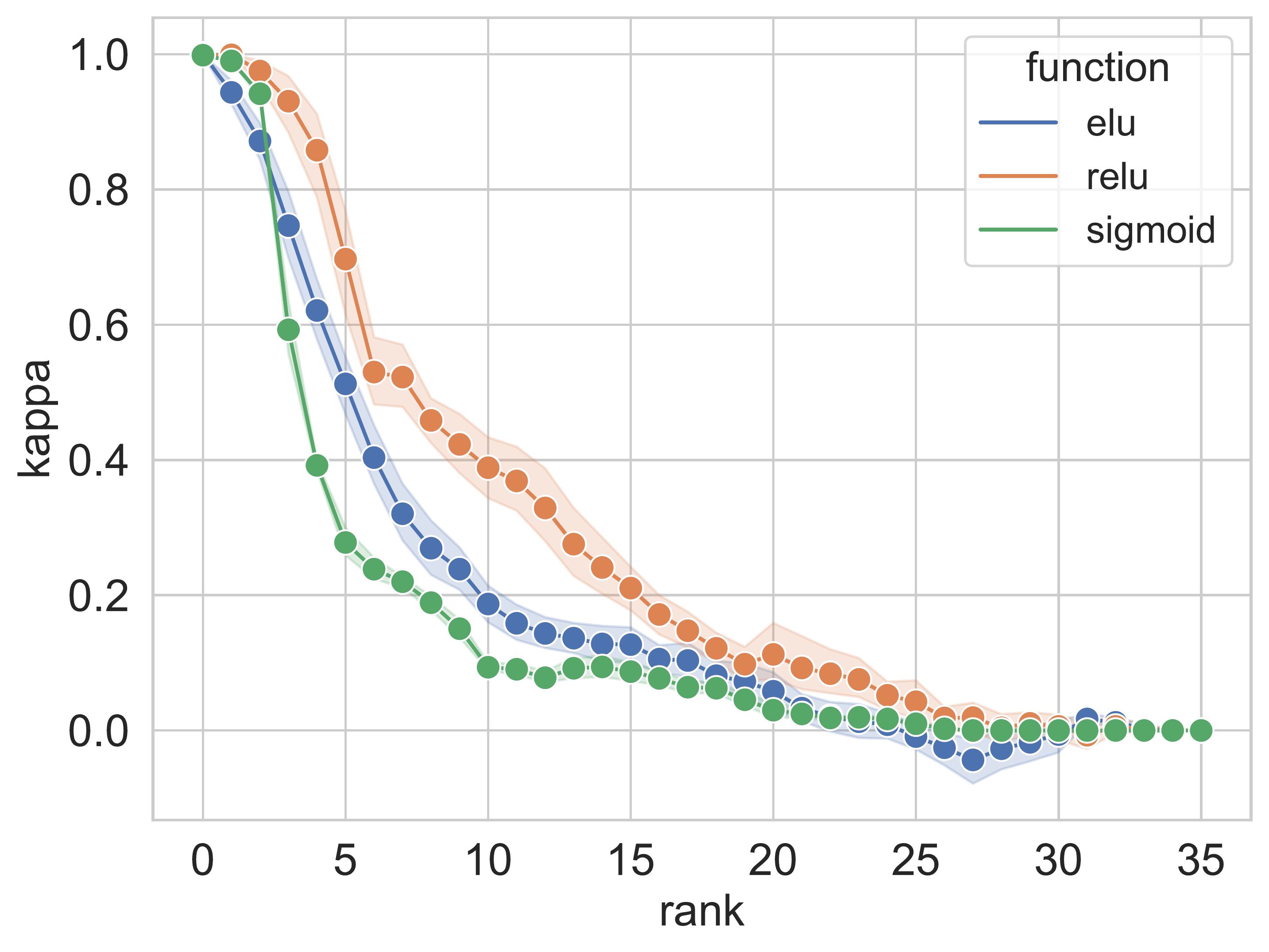}
	\caption{Cardiotocography 3}
	\end{subfigure}

    \begin{subfigure}{0.24\textwidth}
	\center
	\includegraphics[width=\textwidth]{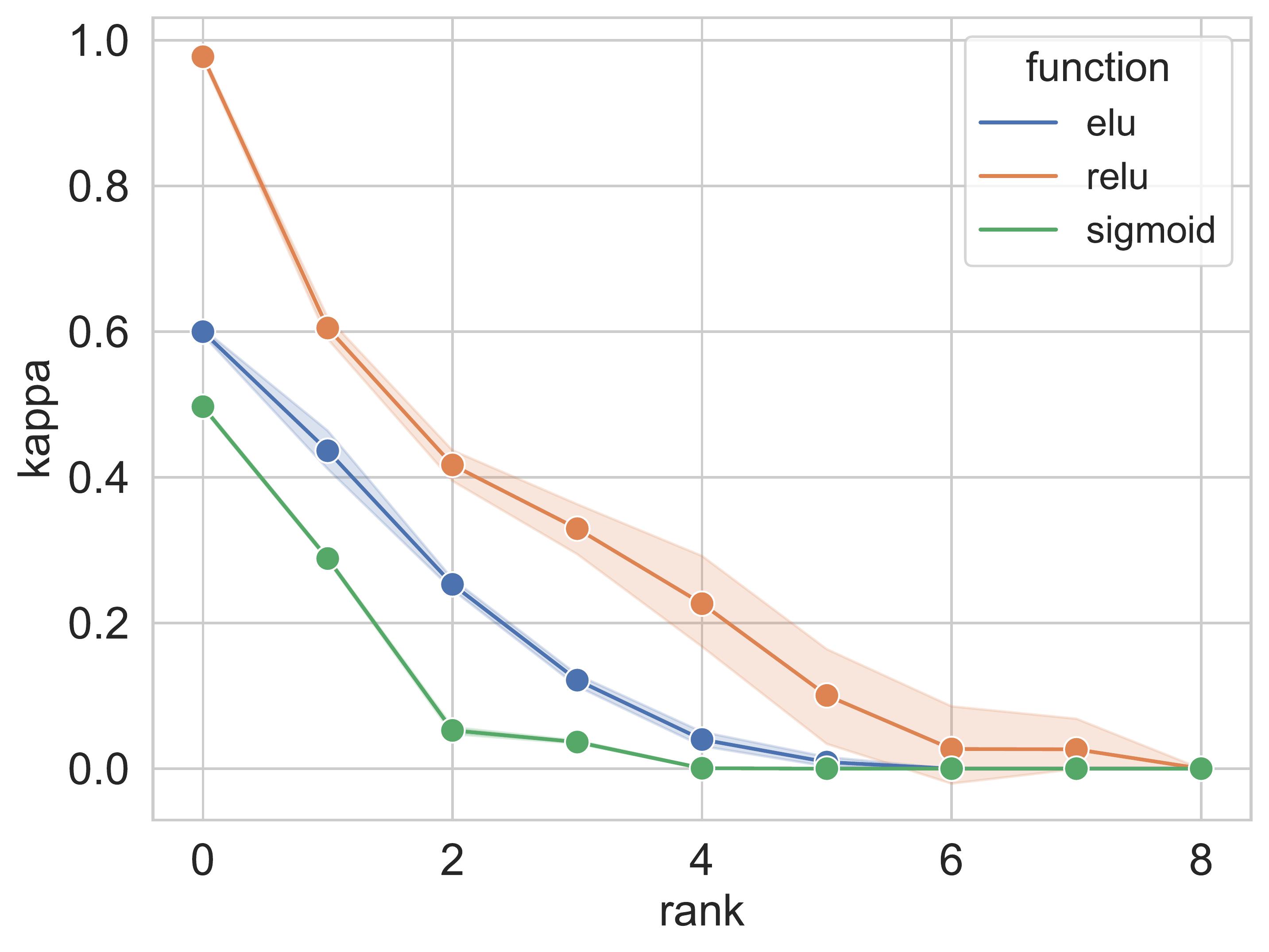}
	\caption{Pima Diabetes}
    \end{subfigure}
    \begin{subfigure}{0.24\textwidth}
	\center
	\includegraphics[width=\textwidth]{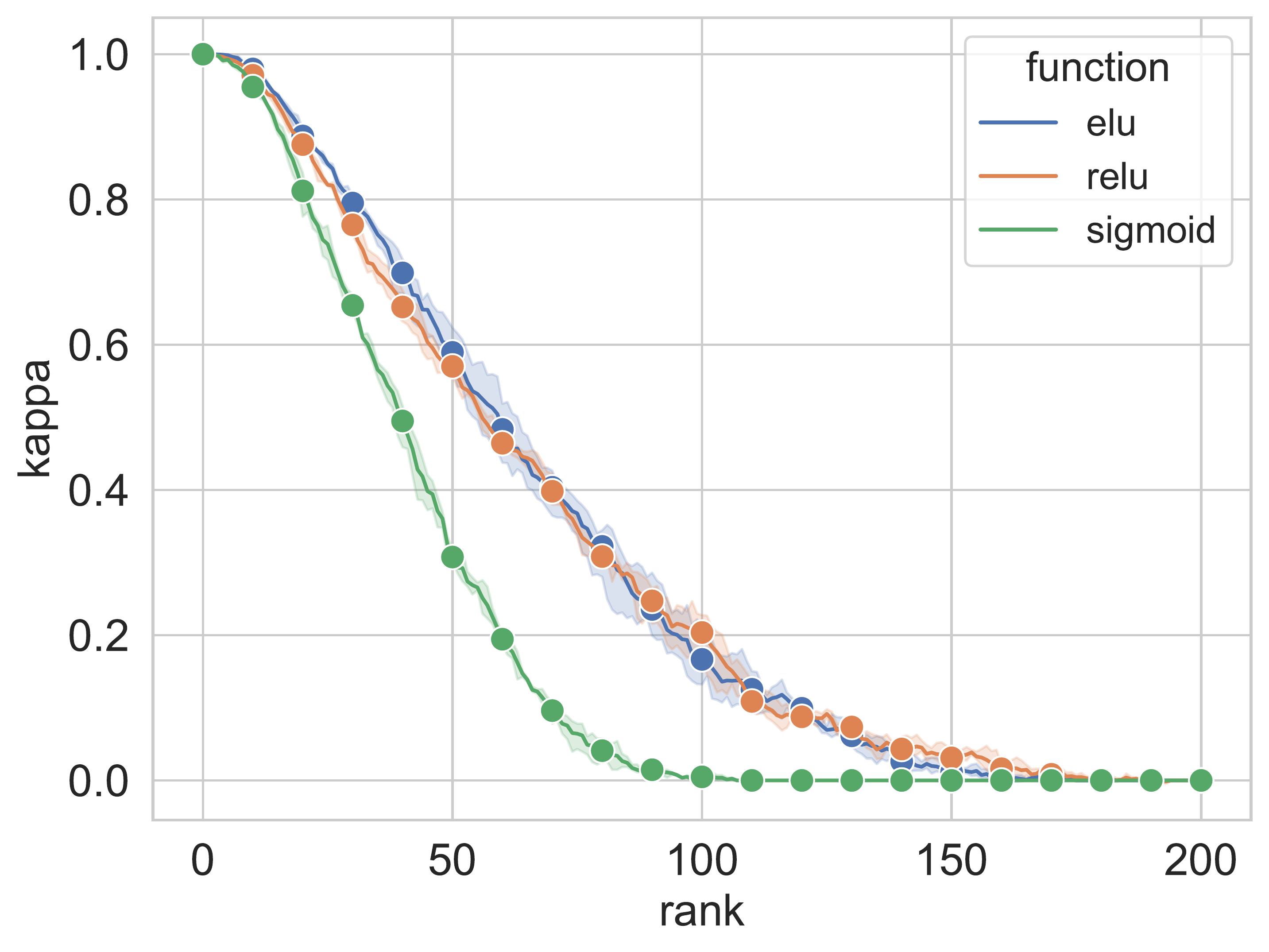}
	\caption{Santander}
	\end{subfigure}
	\begin{subfigure}{0.24\textwidth}
	\center
	\includegraphics[width=\textwidth]{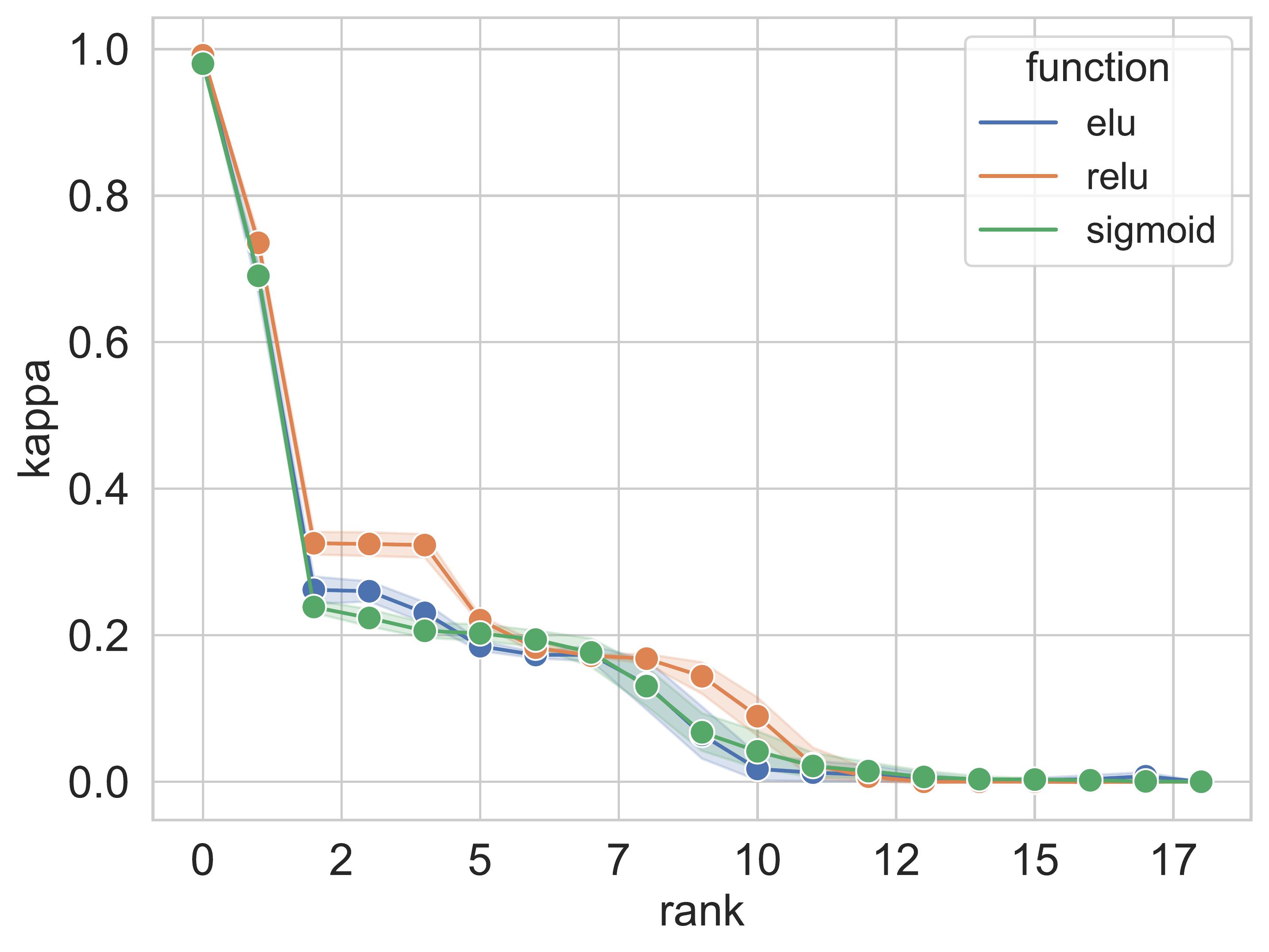}
	\caption{Segment}
	\end{subfigure}
	\begin{subfigure}{0.24\textwidth}
	\center
	\includegraphics[width=\textwidth]{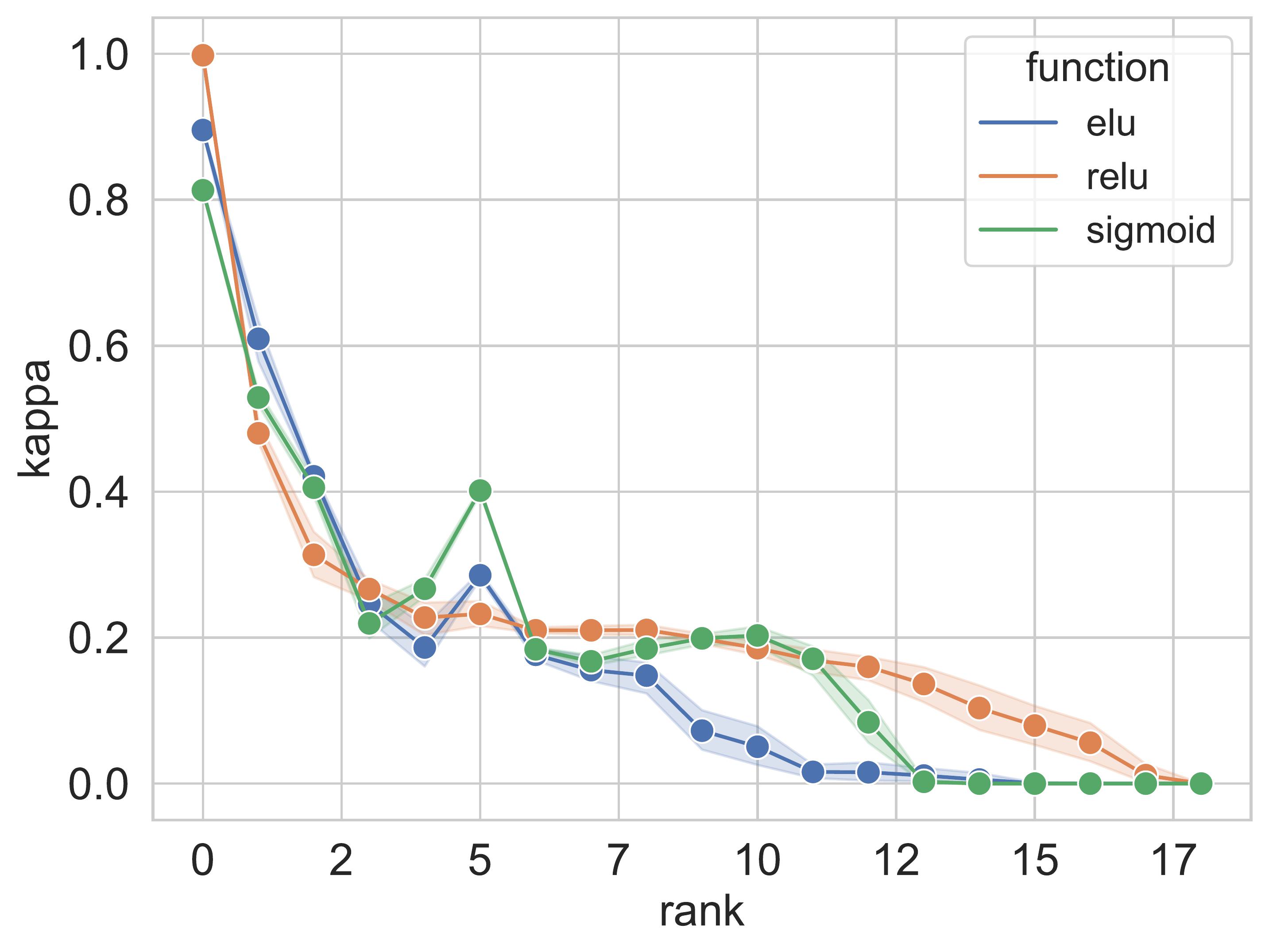}
	\caption{Vehicle}
	\end{subfigure}
	
	\captionsetup{justification=justified}
	\caption{Feature-flipping curves for eight benchmark datasets, considering three different configurations for the activation function of the neural network. The faster the curve goes towards lower values of kappa, the stronger the influence of the features deemed important according to the ranking of composition values.}
\label{fig:curves}
\end{figure*}

Figure \ref{fig:curves} shows the results taking into account three network activation function setups. For all of them, the performance deterioration decreased as the features were modified in the same order they appeared in the feature importance ranking. In other words, altering the most important features as determined by the composition values notably decreased the prediction performance. In the case of high-dimensional datasets, we observed that many features are redundant and produce smaller compositions. This is a common behavior in feature importance methods and is reflected in the slightly slower deterioration of the curves. These results provide reliable evidence in which the composition values reflect the role that each feature plays in a given neuron.

\subsection{Future research directions}
\label{sec:simulations:future}

The promising results reported by the FCP algorithm for the developed case study concerning fairness invite us to consider further research directions. The first one includes expanding the algorithm to other layers and models, such as the Long Short-term Memory, the Gated Recurrent Unit, or Convolutional Neural Networks. Since models are mainly devoted to unstructured data, determining segments or prototypes mapping to structured features would lead to meaningful explanations.

In the context of fairness, the proposed FCP algorithm can be integrated into the backpropagation learning of feed-forward neural networks to mitigate implicit bias. We conjecture that the space described by the relationship bias-accuracy is multimodal, which means that different bias behaviors might be related to the same decision. For example, a neural network trained on biased data can learn to reject a credit application either because the applicant is female or because the applicant (who might be female) has a substantial debt with the bank. The former reason is not ethically acceptable, whereas the latter seems to be reasonable given the bank's financial implications.

Hence, a research direction worthy of investigation concerns the mitigation of either implicit or explicit bias during the training process of feed-forward neural networks devoted to pattern classification. In practice, this can be done by adding a bias term penalizing the decisions that rely on protected features or unprotected ones encoding implicit bias. Equation \eqref{eq:loss-function} depicts an example of a loss function that minimizes the effect of the $k$-th protected composition value on the whole network,

\begin{equation}
\label{eq:loss-function}
\mathcal{L}_k = \underbrace{\sum_{i} \left(\Omega_i(x) - y_{i} \right)^2}_{\text{error term}} + \lambda_1 \underbrace{\sum_{l} \sum_{j} \vartheta_{jk}^{(l)}}_{\text{bias term}} + \lambda_2 \underbrace{\mathcal{R}(W)}_{\text{regularizer}}
\end{equation}

\noindent where

\begin{equation}
\label{eq:regularizer}
\mathcal{R}(W) = \left\Vert W \right\Vert_2^2 = \sum_{l} \sum_{h} \sum_{j} \left( w_{hj}^{(l)} \right)^2
\end{equation}

\noindent such that $y_i$ is the output produced by the $i$-th output neuron, $\Omega(x)$ is the network's output for a given input, $\vartheta_{jk}^{(l)}$ is the $k$-th composition value associated with the $j$-th neuron of the $l$-th layer, while $ w_{hj}^{(l)}$ is the weight arriving at $j$-th neuron. In this loss function, $\lambda_1$ and $\lambda_2$ are positive user-specified parameters to control the importance of the bias and regularization terms, respectively.

Aiming to validate the effectiveness of our approach, it is possible to compute feature attribution scores for protected features when $\lambda_2=0$ and $\lambda_2>0$. In other words, we can study whether the network relies more on protected features when $\lambda_2=0$ than when $\lambda_2>0$. Such an experimental study involving more datasets would provide further evidence of the reliability of the knowledge representations generated by the FCP algorithm.

\section{Concluding remarks}
\label{sec:remarks}

This paper presented an algorithm to describe hidden neurons in feed-forward networks using composition vectors. Such vectors indicate the role of features in each neural processing entity as characterized by the direction and magnitude of composition values. The direction indicates whether the feature excites or inhibits the neuron, while the magnitude quantifies the strength of such impact. An advantage of the FCP algorithm is that it operates using a given instance and the knowledge structures of a previously trained neural network. This means that the explanations it provides help us understand the inner workings of the black box itself instead of only explaining its predictions. At the same time, being model-dependent and relying on structured features can be considered a limitation.

During the experiments, we illustrated the correctness of our algorithm using a case study related to fairness in machine learning in which the ground truth was known. The results using several transfer functions showed that age correlated with the composition values for that feature in the last layer. The results concerning the categorical feature gender indicated alignment between the prediction computed by the network and those obtained from composition values in the last layer. Overall, the composition values allowed us to arrive at the following conclusions: (i) older applicants will likely receive a good credit while younger applicants will likely receive a bad one, and (ii) male applicants are more likely to receive a good credit. These conclusions align with the ground truth, thus confirming the suitability of the FCP algorithm to generate composition-based explanations. An extension of the experiment to benchmark datasets shows that the composition values reflect the role that each feature plays in a given neuron. 

While the proposed algorithm intends to provide meaning to hidden neurons in feed-forward neural networks to explain their inner workings, it paves the way for future research to tackle the bias issue in neural systems. For example, we can modify the backpropagation algorithm to minimize the composition values corresponding to protected features without altering the dataset or significantly affecting accuracy. In that way, the network will be encouraged to make predictions by focusing on unprotected features. Such an improved learning algorithm will be the center of our next research endeavors.

\section*{Acknowledgement}

M. Bello was supported by the funding MCIN/AEI/10. 13039/501100011033/ and FEDER "Una manera de hacer Europa" under grant CONFIA (PID2021-122916NB-I00). Y. Salgueiro is with the Department of Industrial Engineering, Faculty of Engineering, Universidad de Talca, Campus Curic\'o, Chile. Y. Salgueiro was supported by the funding CENIA FB210017, Basal ANID, and the super-computing infrastructure of the NLHPC (ECM-02). A. Jastrzebska’s contribution was funded by the Warsaw University of Technology within the Excellence Initiative: Research University (IDUB) program.

\bibliographystyle{IEEEtran}


\end{document}